\documentclass[runningheads]{llncs}
\usepackage[T1]{fontenc}
\usepackage{graphicx}

\usepackage[accsupp]{axessibility}  %

\usepackage{booktabs}

\RequirePackage{xspace}
\usepackage{upgreek}

\usepackage[labelsep=period]{caption}
\newcommand{\Tref}[1]{Table~\ref{#1}}

\newcommand{\eref}[1]{Eq.~\eqref{#1}}

\newcommand{\fref}[1]{Fig.~\ref{#1}}
\newcommand{\Fref}[1]{Figure~\ref{#1}}
\newcommand{\sref}[1]{Sec.~\ref{#1}}

\newcommand\blfootnote[1]{%
  \begingroup
  \renewcommand\thefootnote{}\footnote{#1}%
  \addtocounter{footnote}{-1}%
  \endgroup
}

\renewcommand{\paragraph}[1]{\noindent \textbf{#1 }}

\usepackage[pagebackref=true,breaklinks=true,colorlinks=true,bookmarks=false]{hyperref} %
\newsavebox\CBox
\def\textBF#1{\sbox\CBox{#1}\resizebox{\wd\CBox}{\ht\CBox}{\textbf{#1}}}

\usepackage{xspace} %
\makeatletter
\DeclareRobustCommand\onedot{\futurelet\@let@token\@onedot}
\def\@onedot{\ifx\@let@token.\else.\null\fi\xspace}

\def\eg{\emph{e.g}\onedot} 
\def\ie{\emph{i.e}\onedot} 
 
\def\etc{\emph{etc}\onedot} 
\def\wrt{w.r.t\onedot} 
\def\etal{\emph{et al}\onedot}
\makeatother

\newcommand{\deformmodule}{deformation module\xspace}
\newcommand{\DeformModule}{Deformation Module\xspace}
\newcommand{\deformgrid}{deformation grid\xspace}
\newcommand{\canonicalmodule}{canonical module\xspace}

\newcommand{\trainingtime}{20 minutes\xspace}
\newcommand{\image}{I}
\newcommand{\viewdir}{\mathbf{d}}
\newcommand{\position}{\mathbf{p}}
\newcommand{\grid}{\mathbf{G}}
\newcommand{\feature}{\mathbf{f}}
\newcommand{\colr}{\mathbf{c}}
\newcommand{\Colr}{\mathbf{C}}
\newcommand{\vieworigin}{\mathbf{o}}
\newcommand{\ray}{\mathbf{r}}

\newcommand{\approximate}[1]{\hat{#1}}

\newcommand{\decay}[1]{\alpha \left( #1 \right)}
\newcommand{\densitysuperscript}{\sigma}
\newcommand{\colorsuperscript}{c}
\newcommand{\deformsuperscript}{d}
\newcommand{\DNet}{F_{\theta_1}^\deformsuperscript}
\newcommand{\CNet}{F_{\theta_2}^\colorsuperscript}
\newcommand{\loss}[1]{\mathcal{L}^{\text{#1}}}
\newcommand{\weight}[2]{w^{\text{#1}}_{\text{#2}}}

\let\oldDelta\Delta
\renewcommand{\Delta}{\rm{\oldDelta}}

\begin{document}
\title{Neural Deformable Voxel Grid for \\ Fast Optimization of Dynamic View Synthesis}
\titlerunning{NDVG}
\author{Xiang Guo\inst{1*} \and
Guanying Chen\inst{2*} \and
Yuchao Dai\inst{1}$^{\dag}$ \and
Xiaoqing Ye\inst{3} \and \\
Jiadai Sun\inst{1} \and
Xiao Tan\inst{3} \and
Errui Ding\inst{3}
}
\authorrunning{X. Guo et al.}
\institute{
Northwestern Polytechnical University, X'an, China \and
FNii and SSE, CUHK-Shenzhen \and
Baidu Inc. \\
\email{\{guoxiang,sunjiadai\}@mail.nwpu.edu.cn},
\email{chenguanying@cuhk.edu.cn},
\email{daiyuchao@nwpu.edu.cn}, 
\email{\{yexiaoqing,dingerrui\}@baidu.com},
\email{tanxchong@gmail.com}\\
}
\maketitle              %

\blfootnote{%
* Authors contributed equally to this work. $^{\dag}$ Yuchao Dai is the corresponding author.}

\begin{abstract}
Recently, Neural Radiance Fields (NeRF) is revolutionizing the task of novel view synthesis (NVS) for its superior performance. In this paper, we propose to synthesize dynamic scenes.
Extending the methods for static scenes to dynamic scenes is not straightforward as both the scene geometry and appearance change over time, especially under monocular setup. Also, the existing dynamic NeRF methods generally require a lengthy per-scene training procedure, where multi-layer perceptrons (MLP) are fitted to model both motions and radiance. In this paper, built on top of the recent advances in voxel-grid optimization, we propose a fast deformable radiance field method to handle dynamic scenes. Our method consists of two modules. The first module adopts a \deformgrid to store 3D dynamic features, and a light-weight MLP for decoding the deformation that maps a 3D point in the observation space to the canonical space using the interpolated features. The second module contains a density and a color grid to model the geometry and density of the scene. The occlusion is explicitly modeled to further improve the rendering quality. Experimental results show that our method achieves comparable performance to D-NeRF using only \trainingtime for training, which is more than $70\times$ faster than D-NeRF, clearly demonstrating the efficiency of our proposed method.

\end{abstract}

\section{Introduction}
Novel view synthesis (NVS) is a long-standing problem in computer vision and graphics, and has many applications in augmented reality, virtual reality, content creation, \etc. Recently, neural rendering methods have achieved significant progress in this problem~\cite{mildenhall2020_nerf_eccv20,yariv2020multiview,niemeyer2020differentiable}.
In particular, the neural radiance fields (NeRF)~\cite{mildenhall2020_nerf_eccv20} produces photorealistic rendering by representing a static scene with a multi-layer perception (MLP), which maps a 5D input (3D coordinate and 2D view direction) to its density and color. Recently, a series of works extend NeRF based framework from static scenes to dynamic scenes~\cite{gao2021dynamic_iccv21,li2021_nsff_cvpr21,xian2021_space_cvpr21,tretschk2020_nonrigid_iccv21,park2021_nerfies_iccv21,pumarola2021_dnerf_cvpr21,wang2021_dctnerf_arxiv,du2021_nerflow_iccv21}.

Novel view synthesis of a dynamic scene from a monocular video is still a very challenging problem. Besides the difficulties to recover motions and geometries with only one observation at each time step, the training process usually takes days which hinders applications in practice.
The NeRF based methods, at each iteration, require millions of network queries to obtain colors and densities of the sampled points for the sampled rays, based on which volume rendering computes the pixel colors~\cite{kajiya1984ray}. In dynamic condition, the methods are even more complex with deformation model, \eg, D-NeRF~\cite{pumarola2021_dnerf_cvpr21} optimizes a large deformation network and a canonical network to fit a dynamic scene, requiring more than $27$ hours to converge. How to develop an efficient and accurate dynamic view synthesis method remains an open problem. 

In a static scenario, to reduce the training time for a scene, some methods propose first to train the model on a dataset consisting of multiple scenes~\cite{chen2021mvsnerf,yu2021_pixelnerf_cvpr21,trevithick2021grf,wang2021_ibrnet_cvpr21}, and then finetune it on the target scene, reducing the optimization time to several minutes.
However, these methods rely on a large training dataset and a lengthy pre-training time.

\begin{figure}[t] \centering
    \includegraphics[width=\textwidth]{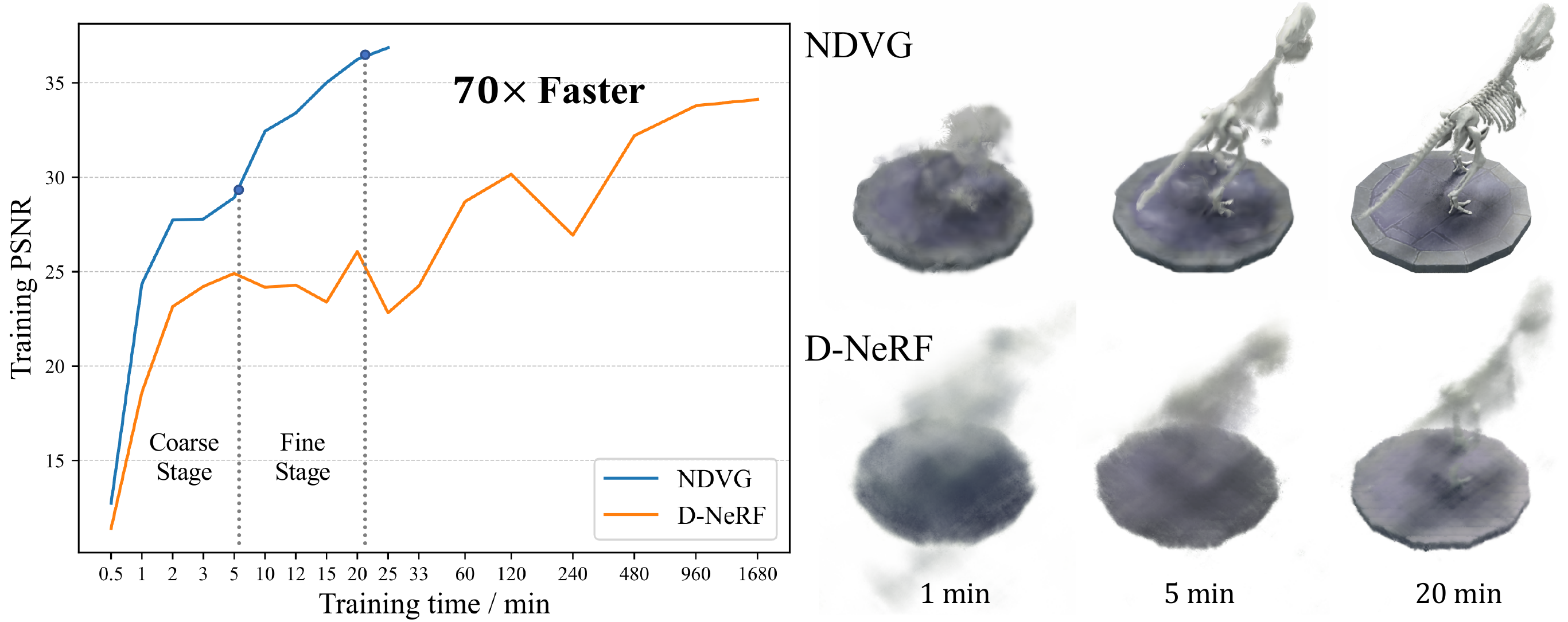}
    \caption{\textbf{Neural Deformable Voxel Grid~(NDVG)} for fast optimization of dynamic view synthesis. Left side of the figure shows that our method achieves a super fast convergence in $20$ minutes, which is $70\times$ faster than the D-NeRF method. Right side of the figure visualizes the results after training with $1$, $5$ and $20$ minutes.} \label{fig:teaser}
\end{figure}

Very recently, the voxel-grid representation has been exploited to speed up the optimization of radiance fields~\cite{yu2021_plenoctrees_iccv21,sun2021direct,muller2022instant}. These methods are able to optimize a scene representation from scratch within just a few minutes, significantly accelerating the training speed without any pre-training. The key idea is to replace the time-consuming deep network query with the fast trilinear voxel-grid interpolation. However, these methods are tailored for static scenes and cannot be directly applied to handle dynamic scenes. 

To the best of our knowledge, there are few works for fast optimization dynamic NeRF. Also, it is challenging to design a compact deformation model, while having enough capacity. In this paper, we propose a fast optimization method, named NDVG for dynamic scene view synthesis based on the voxel-grid representation.
Our method consists of a \deformmodule and a \canonicalmodule. The \deformmodule maps a 3D point in the observation space to canonical space, and volume rendering is performed in the canonical space to render the pixel color and compute image reconstruction loss.
In contrast to static scene applications where the occlusion is only determined by the viewing direction when the scene is known, the occlusion in the dynamic scene NeRF is determined by both view direction and motion (or view time) and should be taken well care of. Hence, we designed a particular occlusion handling module to explicitly model the occlusion to further improve the rendering quality. 

In summary, the key contributions of this paper are as follows:
\begin{itemize}
    \item[$\bullet$] We propose a fast deformable radiance field method based on the voxel-grid representation to enable space-time view synthesis for dynamic scenes. To the best of our knowledge, this is the first method that integrates the voxel-grid optimization with deformable radiance field.
    \item[$\bullet$] We introduce a deformation grid to store the 3D dynamic features and adopt a light-weight MLP to decode the feature to deformation. Our method explicitly models occlusion to improve the results.
    \item[$\bullet$] Our method produces rendering results comparable to D-NeRF \cite{pumarola2021_dnerf_cvpr21} within only \trainingtime, which is more than $70\times$ faster (see~\fref{fig:teaser}).
\end{itemize}

\section{Related Work}
\paragraph{Novel View Synthesis} 
Rendering a scene from arbitrary views has a long history in both vision and graphics~\cite{buehler2001_unstructured,chen1993view,levoy1996light,greene1986environment}, and surveys of recent methods can be found in~\cite{shum2000review,tewari2020state,tewari2021advances}.
Traditional methods need to explicitly build a 3D model for the scene, such as point clouds~\cite{aliev2020_neural_eccv20} or meshes~\cite{riegler2020_freenvs_eccv20,riegler2021_stablenvs_cvpr21,thies2019_deferred_tog19,hedman2018_deepibr_tog18}, and then render a novel view from this geometry. 
Another category of methods explicitly estimates depth and then uses it to warp pixels or learned feature to a novel view, such as ~\cite{kalantari2016learning_tog16,penner2017_soft_tog17,choi2019_extreme_nvs_iccv19,riegler2020_freenvs_eccv20,riegler2021_stablenvs_cvpr21,flynn2016_deepstereo_cvpr16,xu2019_deep_tog19}.
Numerous other works using multi-plane images (MPIs) to represent scenes~\cite{flynn2019_deepview_cvpr19,mildenhall2019_local_tog19,srinivasan2020_lighthouse_cvpr20,srinivasan2019_pushing_cvpr19,zhou2018_stereo_tog18,tucker2020single_cvpr20,huang2020semantic_eccv20,habtegebrial2020_generative_nips20}, but the MPIs representation can only support relatively limited viewpoint changes during inference.

\paragraph{Neural Scene Representation} 
Recently, neural scene representations dominate novel view synthesis.
In particular, Mildenhall~\etal propose NeRF~\cite{mildenhall2020_nerf_eccv20} to use MLPs to model a 5D radiance field, which can render impressive view synthesis for static scenes captured.
Since then, many follow-up methods have extended the capabilities of NeRF, including relighting~\cite{boss2021nerd,srinivasan2021nerv,zhang2021nerfactor}, handling in-the-wild scenarios~\cite{martin2021_nerfw_cvpr21}, extending to large unbounded 360$^{\circ}$ scenes~\cite{Zhang20arxiv_nerf++}, removing the requirement for pose estimation~\cite{wang2021_nerfminus_arix,lin2021barf}, incorporating anti-aliasing for multi-scale rendering~\cite{barron2021mipnerf}, and estimating the 6-DoF camera poses~\cite{yen2021inerf}.

\paragraph{Fast NeRF Rendering and Optimization}
Rendering and optimization in NeRF-like schemes are very time-consuming, as they require multiple samples along each ray for color accumulation. 
To speed up the rendering procedure, some methods predict the depth or sampling near the surface to guide more efficient sampling~\cite{neff2021_donerf_egsr21,piala2021terminerf}. Other methods use the octree or sparse voxel grid to avoid sampling points in empty spaces~\cite{liu2020_nsvf_nips20,yu2021_plenoctrees_iccv21,lombardi2021mixture}.
In addition, some methods subdivided the 3D volume into multiple cells that can be processed more efficiently, such as DeRF~\cite{rebain2021_derf_cvpr21} and KiloNeRF~\cite{Reiser2021_kiloNeRF_iccv21}.
AutoInt~\cite{lindell2021_autoint_cvpr21} reduces the number of evaluations along a ray by learning partial integrals.
However, these methods still need to optimize a deep implicit model, leading to a lengthy training time.
To accelerate the optimization time on a test scene, some methods first train the model on a large dataset to learn a scene prior, and then finetune it on the target scene~\cite{chen2021mvsnerf,trevithick2021grf,yu2021_pixelnerf_cvpr21,wang2021_ibrnet_cvpr21}. However, these methods require time-consuming pretraining.

More recently, the voxel-grid representation has been exploited to speed up the optimization of radiance field~\cite{yu2021plenoxels,sun2021direct,muller2022instant}. Plenoxels~\cite{yu2021plenoxels} represents a scene as a 3D grid with spherical harmonics, which can be optimized from calibrated images via gradient methods and regularization without any neural components. Similarly, DVGO~\cite{sun2021direct} optimizes a hybrid explicit-implicit representation that consists of a dense grid and a light-weight MLP.
Although these methods achieve a fast optimization speed, they are only applicable to static scenes and thus cannot be used to render dynamic scenes.

\paragraph{Dynamic Scene Modeling}
Recently, several concurrent methods have extended NeRF to deal with dynamic scenes~\cite{gao2021dynamic_iccv21,li2021_nsff_cvpr21,xian2021_space_cvpr21,tretschk2020_nonrigid_iccv21,park2021_nerfies_iccv21,pumarola2021_dnerf_cvpr21,wang2021_dctnerf_arxiv,du2021_nerflow_iccv21}.

NeRFlow~\cite{du2021_nerflow_iccv21} learns a 4D spatial-temporal representation of a dynamic scene from a set of RGB images. 
Yoon~\etal~\cite{yoon2020_nvidiadataset_cvpr20} propose to use an underlying 4D reconstruction, combining single-view depth and depth from multi-view stereo to render virtual views with 3D warping. 
Gao~\etal~\cite{gao2021dynamic_iccv21} jointly train a time-invariant model (static) and a time-varying model (dynamic), and regularize the dynamic NeRF by scene flow estimation, finally blending the results in an unsupervised manner.
NSFF~\cite{li2021_nsff_cvpr21} models the dynamic scene as a time-variant continuous function of appearance, geometry, and 3D scene motion.
DCT-NeRF~\cite{wang2021_dctnerf_arxiv} uses the Discrete Cosine Transform (DCT) to capture the dynamic motion and learn smooth and stable trajectories over time for each point in space. 

D-NeRF~\cite{pumarola2021_dnerf_cvpr21}, NR-NeRF~\cite{tretschk2020_nonrigid_iccv21} and Nerfies~\cite{park2021_nerfies_iccv21} first learn a static canonical radiance field for capturing geometry and appearance, and then learn the deformation/displacement field of the scene at each time instant \wrt the canonical space. Xian \etal~\cite{xian2021_space_cvpr21} represent a 4D space-time irradiance field as a function that maps a spatial-temporal location to the emitted color and volume density. 

Although promising results have been shown for dynamic view synthesis, these methods all require a long optimization time to fit a dynamic scene, limiting their wider applications.

As the concurrent research of our work, there are few works which aim to speed up training of dynamic NeRF \cite{fang2022_TANV_arxiv,gan2022_V4D_arxiv,liu2022_DeVRF_arxiv}. TiNeuVox \cite{fang2022_TANV_arxiv} uses a small MLP to model the deformation and uses the multi-distance interpolation to get the feature for radiance network which estimates the density and color. Compared to TiNeuVox \cite{fang2022_TANV_arxiv}, we propose a deformation feature grid to enhance the capability of the small deformation network and not effect training speed. V4D \cite{gan2022_V4D_arxiv} uses the 3D feature voxel to model the 4D radiance field with additional time dimension concatenated and proposes look-up tables for pixel-level refinement. While V4D~\cite{gan2022_V4D_arxiv} focuses mainly on improving image quality, the speed up of training is not significant compared with TiNeuVox \cite{fang2022_TANV_arxiv} and ours. DeVRF \cite{liu2022_DeVRF_arxiv} also builds on voxel-grid representation, which proposes to use multi-view data to overcome the nontrivial problem of the monocular setup. Multi-view data eases the learning of motion and geometry compared with ours which uses monocular images.

\section{Neural Deformable Voxel Grid}
Given an image sequence $\{\image_t\}_{t=1}^{T}$ with camera poses $\{{\rm{\mathbf{T}}}_t\}_{t=1}^{T}$ of a dynamic scene captured by a monocular camera, our goal is to develop a fast optimization method based on the radiance fields to represent this dynamic scene and support novel view synthesis at different times.

\subsection{Overview}
Mathematically, given a query 3D point $\position$, the view direction $\viewdir$, and a time instance $t$, we need to estimate the corresponding density $\sigma$ and color $\colr$.

A straightforward way is to directly use an MLP to learn the mapping from $(\position, \viewdir, t)$ to $(\sigma, \colr)$. However, existing dynamic methods show that such a high-dimensional mapping is difficult to learn, and they propose a framework based on the canonical scene to ease learning difficulty~\cite{pumarola2021_dnerf_cvpr21,tretschk2020_nonrigid_iccv21,park2021hypernerf}.
These methods adopt a deformation MLP to map a 3D point in the observation space to a static canonical space as $\Uppsi_t: (\position, t) \to \Delta \position $. The density and color are then estimated in the canonical space as $\Uppsi_p: (\sigma, \colr) = f(\position + \Delta \position, \viewdir)$.
However, to render the network with the capability of handling complex motion, large MLPs are inevitable in existing methods and therefore result in a long optimization time (\ie, from hours to days).

\begin{figure}[t] \centering
    \includegraphics[width=\textwidth]{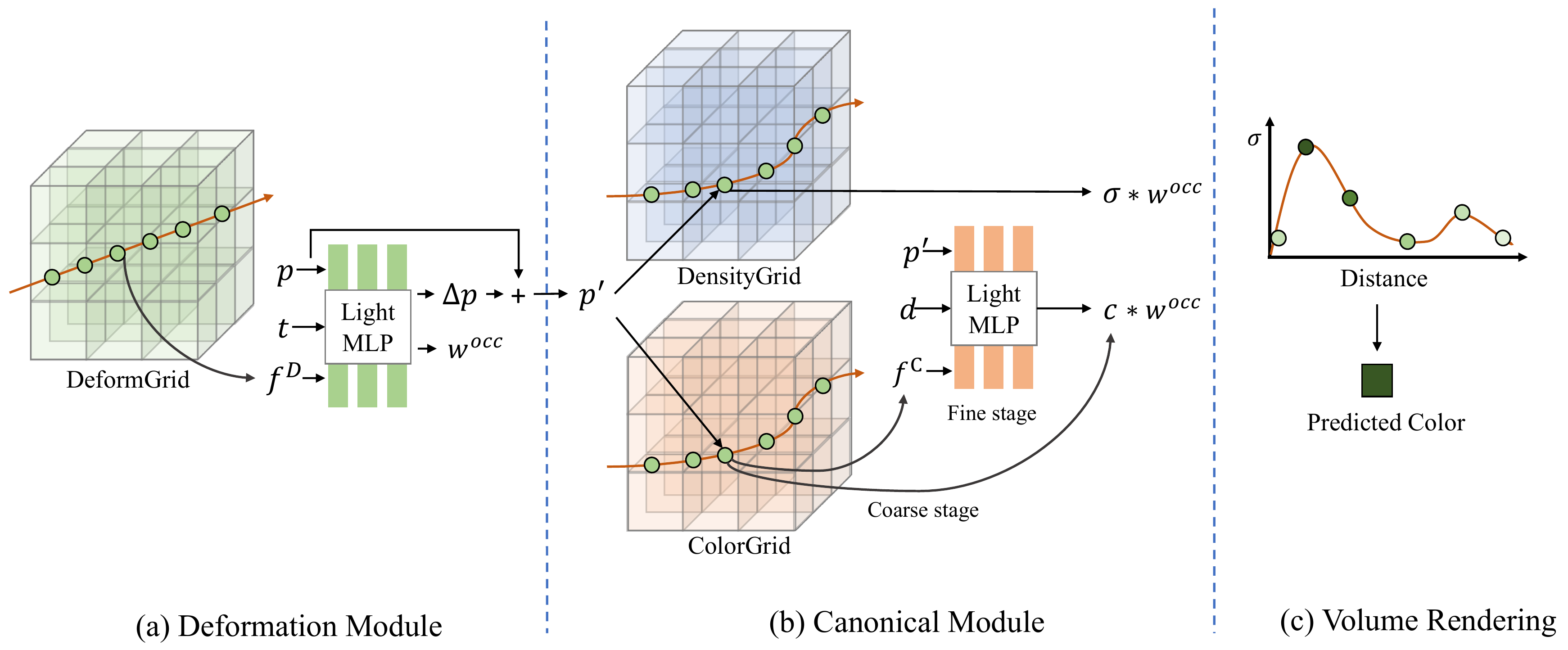}
    \caption{\textbf{Overview of our proposed method.} Our method consists of (a) a \emph{\deformmodule} to model the motion of the space points and (b) a \emph{\canonicalmodule} to model the radiance field of the static scene at the canonical time.
    To render a ray shooting from the camera center, we compute the deformation of all sampled points and transform the sampled points to the canonical space, where the density and color are computed. The pixel color can then be rendered by (c) volume rendering.} \label{fig:pipeline}
\end{figure}

Motivated by the recent successes of voxel-grid optimization in accelerating the training of static radiance field~\cite{sun2021direct,yu2021plenoxels}, we introduce the voxel-grid representation into the canonical scene representation based framework to enable fast dynamic scene optimization.
Our method consists of a \emph{\deformmodule} and a \emph{\canonicalmodule} (see~\fref{fig:pipeline}).
The key idea is to replace most of the heavy MLP computations with the fast voxel-grid feature interpolation.

\subsection{\DeformModule for Motion Modeling} \label{sec:deform_module}
Assuming the canonical space is at time $t_{can}$, the \deformmodule estimates the offset $\Delta\position$ of a point $\position$ at any time $t$ to the canonical space.
As the input has 4 dimensions (\ie, 3 for $\position$ and 1 for $t$), it is inefficient to directly store the offsets in a 4D feature grid. 
Therefore, we adopt a hybrid explicit-implicit representation that consists of a 3D feature grid and a light-weight MLP to decode the interpolated feature.

\paragraph{Neural Deformable Voxel Grid (NDVG)}
We first build a 3D deformation feature grid $\grid^\deformsuperscript \in \mathbb{R}^{N_x\times N_y\times N_z\times N_c}$ with a resolution of $N_x\times N_y\times N_z$, and the feature vector in each voxel has a dimension of $N_c$. For a continuous 3D coordinate $\position$, its deformation feature can be quickly queried from the deformation feature grid using trilinear interpolation:
\begin{align}
 \text{interp}(\position, \grid^\deformsuperscript): (\mathbb{R}^3, \mathbb{R}^{N_x\times N_y\times N_z\times N_c}) \xrightarrow[]{} \mathbb{R}^{N_c}.
\end{align}

To obtain the offset from the observation space to the canonical space for a query point, we introduce a light-weight MLP $\DNet$. It takes the coordinate $\position$, time $t$, and the interpolated feature as input, and regresses the offset $\Delta\position$.
\begin{align}
    \feature^\deformsuperscript = \text{interp}(\position, \grid^\deformsuperscript), 
    \quad
    \Delta\position =  
    \begin{cases} 
        \DNet(\position, t, \feature^\deformsuperscript) \quad \text{if $t \neq t_{can}$,} \\
        0 \qquad\qquad\quad \ \ \text{otherwise.}
    \end{cases}
   \label{eq:deformation}
\end{align}

Following~\cite{mildenhall2020_nerf_eccv20}, we apply positional encoding for $\position$ and $t$. 
Finally, we get the position of $\position$ in the canonical space $\position'$ as: $\position' = \position + \Delta\position$.

The deformation feature grid can provide learnable features that encode the deformation information of the points in the 3D space, so that a light-weight MLP is capable to model the deformation of the 3D space.

\subsection{Canonical Module for View Synthesis} \label{sub:canonical}
To render the pixel color of a camera ray in the observation space, we first sample $K$ points on the ray and transform them to the canonical space via our \deformmodule. The \canonicalmodule then computes the corresponding densities and colors for volume rendering to composite the pixel color.

\paragraph{Density and Color Grids}
The \canonicalmodule contains a density grid to model the scene geometry, and a hybrid explicit-implicit representation (contains a color feature grid and a light-weight MLP) to model the view-dependent effect.

The density grid $\grid^\densitysuperscript \in \mathbb{R}^{N'_x\times N'_y\times N'_z}$ with a resolution of $N'_x\times N'_y\times N'_z$ stores the density information of the scene. 
Following~\cite{sun2021direct}, for a 3D query point $\position'$,  we perform trilinear interpolation on the grid and apply post-activation with the softplus activation function to get the density: $\sigma = \text{softplus}(\text{interp}(\position', \grid^\densitysuperscript))$

The color feature grid $\grid^\colorsuperscript \in \mathbb{R}^{H'\times W'\times D'\times N_c'}$ with a resolution of $H'\times W' \times D'$ stores the color feature of the scene, where the feature dimension is $N_c'$. To obtain the view dependent $\colr$ for a 3D point $\position'$, we use a light-weight MLP $\CNet$ to decode the interpolated color feature $\feature^\colorsuperscript$. Positional encoding is applied on $\position'$ and $\viewdir$.

\begin{align}
    \feature^\colorsuperscript = \text{interp}(\position', \grid^\colorsuperscript), \quad \colr = \CNet( \position', \viewdir, \feature^\colorsuperscript).
\end{align}

\subsection{Occlusion-aware Volume Rendering}
\label{sec:occlusion}
Volume rendering is the key in radiance fields based methods to render pixel color in a differentiable manner.
For a ray $\ray(w) = \vieworigin + w \viewdir$ emitted from the camera center $\vieworigin$ with view direction $\viewdir$ through a given pixel on the image plane, the estimated color $\approximate{\Colr}(\ray)$ of this ray is computed as
\begin{gather} %
    \approximate{\Colr}(\ray) =  
        \sum_{k=1}^{K} T(w_k) \, \decay{\sigma(w_k) \delta_{k}} \, \colr(w_k) \, , 
    \quad
    T(w_k) = 
        \exp \left(-\sum_{j=1}^{k-1} \sigma(t_{j}) \delta_{j} \right) \, ,\label{eqn:nerf2}
\end{gather}
where $K$ is the sampled points in the ray, $ \delta_{k}$ is the distance between adjacent samples on the ray, and $\decay{\sigma(w_k) \delta_{k}} = 1 - \exp (-\sigma(w_k) \delta_{k})$. 

\begin{figure}[t] \centering
    \includegraphics[width=0.9\textwidth]{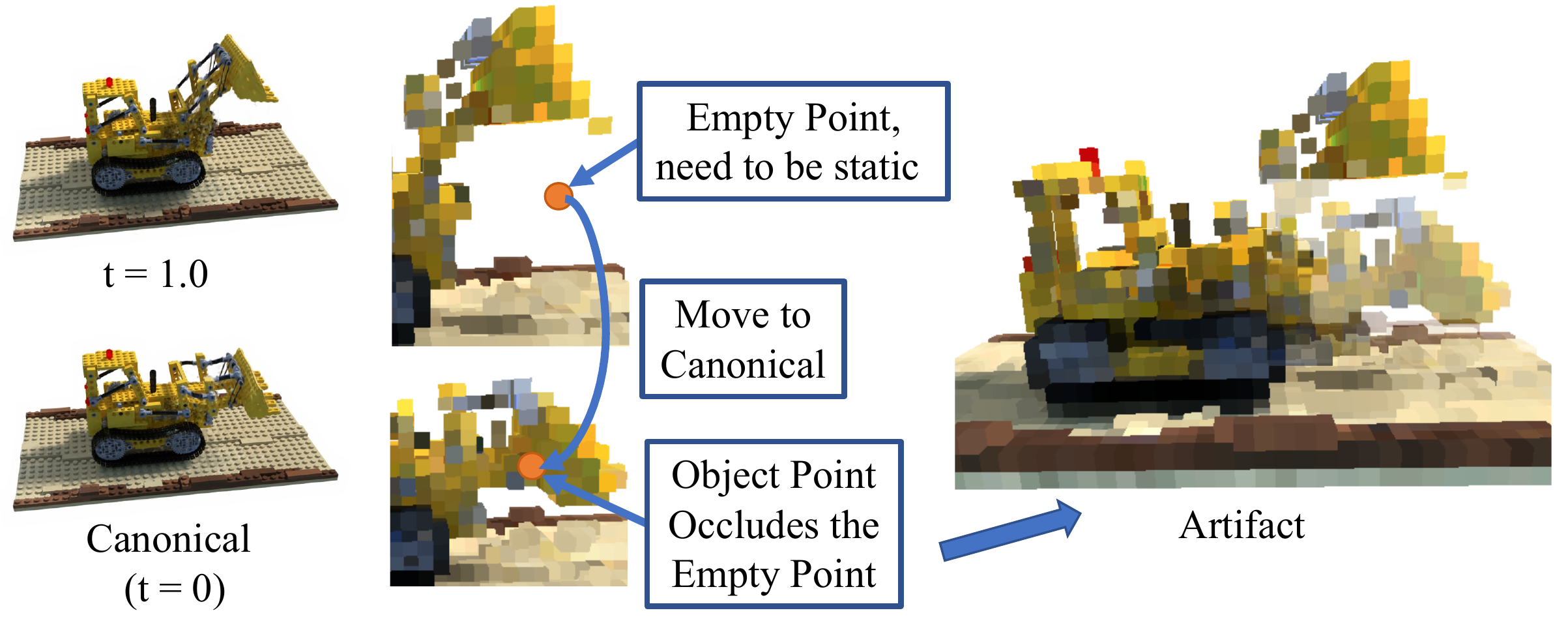}
    \caption{\textbf{Effect of Occlusion.} If the static empty point in current time is occluded by the object point in canonical time, this empty point will bring the object point back to current time, which creates artifacts.} \label{fig:occlusion_lite}
\end{figure}

\paragraph{Occlusion Problem}
Note that for dynamic scenes, the occlusion will cause problems in the canonical scene-based methods.
We assume the ``empty'' points (\ie, non-object points) in the space are static.
If a 3D point is the empty point at time $t$ but occupied by an object point in the canonical space, the occlusion happens. 
This is because the ideal deformation for an empty point in the observation space is zero, then this point will be mapped to the same location (but occupied by an object point) in the canonical space (see~\fref{fig:occlusion_lite} for illustration). 
As a result, the obtained density and color value for this empty point will be non-zero, leading to incorrect rendering color.

\paragraph{Occlusion Reasoning}
To tackle the occlusion problem, we additionally estimate an occlusion mask for a query point using the deformation MLP $\DNet$. Then \eref{eq:deformation} is revised as:
\begin{equation}
    (\Delta{\position}, {\it w^{occ}}) 
    = {\it \DNet}(\position, {\it t}, \feature^D). \label{eq:occlusion}
\end{equation} 

If a query point is an empty point at time $t$ but is occupied by an object point in the canonical space, the estimated $w^{occ}$ should be $0$.
Based on the estimated occlusion mask $w^{occ}$, we filter the estimated density and color before sending them to volume rendering to remove the influence of the occluded points:
\begin{align}
    \sigma' = \sigma \times w^{occ}, \quad \colr' = \colr \times w^{occ}.
\end{align}
The occlusion estimation is optimized by the image reconstruction loss in an end-to-end manner. By taking occlusion into account, our method is able to achieve better rendering quality.

\section{Model Optimization}
There is a trade-off between grid resolution and time consumption. Using a higher grid resolution generally leads to better image quality but at the cost of a higher computational time. Meanwhile, the space is dominated by empty space, which takes no effect on the synthesized images. 

\paragraph{Loss Function}
We utilize a series of losses to optimize the model. 
The first is the photometric loss, which is the mean square error (MSE) between the estimated ray color $\approximate{\Colr}(\ray)$ and the ground-truth color $\Colr(\ray)$ as
\begin{gather}
    \loss{photo}=\frac{1}{|\mathcal{R}|} \sum_{r\in\mathcal{R}} \left\| \approximate{\Colr}(\ray) - \Colr({\ray}) \right\|_2^2,
\end{gather}
where $\mathcal{R}$ is the set of rays sampled in one batch.

Also, similarly to \cite{sun2021direct}, we use the sampled point color supervision $\mathcal{L}^{\mathrm{ptc}}$, which optimizes the color of sampled points with top $N$ weights that contribute most to the rendered color of the ray. In addition, we use the background entropy loss $\loss{bg}$ to encourage the points to concentrate on either foreground or background. 

For the \deformmodule, we apply the $L_1$ norm regularization to the estimated deformation based on the prior that most points are static:
\begin{gather}
    \loss{d\_norm}=  \frac{1}{|\mathcal{R}|} \sum_{r\in\mathcal{R}} \sum_{i=1}^{K} \left\| \Delta\position_i \right\|_1.
\end{gather}

Moreover, we apply the total variation loss on deformation feature grid to smooth the voxel features as 
\begin{gather}
    \loss{d\_tv}=\frac{1}{|\mathcal{V}|}
\sum_{\substack{\mathbf{v} \in \mathcal{V} c \in [N_c]}}
    \sqrt{\Delta_x^2(\mathbf{v},c) + \Delta_y^2(\mathbf{v},c) + \Delta_z^2(\mathbf{v},c)},
\end{gather}
where $\Delta_x^2(\mathbf{v},c)$ represents the square difference between $c$-th element in voxel $\mathcal{V}(i,j,k)$ and voxel $\mathcal{V}(i+1,j,k)$, and the same for $\Delta_y^2(\mathbf{v},c)$ and $\Delta_z^2(\mathbf{v},c)$.

The overall loss function for can be written as
\begin{gather}
\loss{} =\loss{photo}+\weight{ptc}{}\!\cdot\!\loss{ptc} + \weight{bg}{}\!\cdot\!\loss{bg} + \weight{d\_norm}{}\!\cdot\!\loss{d\_norm} + \weight{d\_tv}{}\!\cdot\!\loss{d\_tv},
\end{gather}
where $\weight{ptc}{}$, $\weight{bg}{}$, $\weight{df\_norm}{}$, and $\weight{df\_tv}{}$ are weights to balance each components in the final coarse loss.

Following \cite{sun2021direct}, we use the same strategy for voxel allocation, model point sampling, and low-density initialization. Considering that the deformation is more sensitive to resolution compared with radiance (density and color), we set a higher grid resolution to the \deformmodule than the \canonicalmodule. 

\paragraph{Progressive Training}
Typically, when a time step goes further away from the canonical time, the deformation between this time step and the canonical time step get larger. To reduce the learning difficulty, we use a progressive training strategy. 
The optimization starts with training images close to canonical time, and progressively adds images with further time steps.

\paragraph{Coarse-to-fine Optimization} To speed up the training while maintaining the model capacity, we adopt a coarse-to-fine optimization procedure following~\cite{mildenhall2020_nerf_eccv20,sun2021direct}. We first optimize a coarse model to roughly recover the deformation and the canonical space geometry. Then, we use the coarse model to locate the object region and filter out a large portion of empty space, after which a fine model is optimized to recover a more accurate and detailed deformation and geometry. In the coarse model, we do not model the view-dependent effect, and the feature dimension of color grid $K'$ is set to $3$, which directly corresponds to the RGB color. Based on the optimized coarse module, we apply empty space filtering strategies, including finding fine-stage bounding box and empty point filtering to speed up training. Also, we initialize the fine-stage model with model weight trained during coarse stage. We present details of our empty space filtering strategies and fine model design in the supplementary material.

\section{Experiments}

\subsection{Dataset and Metrics}

We evaluate our method on the D-NeRF dataset~\cite{pumarola2021_dnerf_cvpr21}, which contains eight dynamic scenes with 360$^\circ$ viewpoint settings. Beside synthetic dataset, we also conduct experiments on real scenes, proposed by HyperNeRF \cite{park2021hypernerf}. This dataset captures images of real dynamic scenes with a multi-view camera rig consisting of 2 phones with around a 16cm baseline. The dynamic motion consists of both rigid and non-rigid deformation with unbounded scenes.
We use several metrics for the evaluation: Peak Signal-to-Noise Ratio (PSNR), Structural Similarity (SSIM), and Learned Perceptual Image Patch Similarity (LPIPS)~\cite{zhang2018_lpips_CVPR}. 

\subsection{Implementation Details}
We set the expected voxel number to 1,664k and $190^3$ for the grid in Deformation Module in the coarse and fine stages, respectively. For Canonical Module, we set the expected voxel number to 1,024k and $160^3$. 
The light-weight MLP in the \deformmodule has 4 layers, each with a width of 64. 
The light-weight MLP in the \canonicalmodule has 3 layers, each with a width of 128. 

When training with full resolution images, we train 10k and 20k iterations for the coarse and fine stages for all scenes. 
When training with half-resolution images, we reduce the iteration to 5k and 10k for coarse and fine stages. %
In terms of positional encoding, we set the frequency to 5 for position and time, and 4 for direction.
We use the Adam optimizer~\cite{KingmaB15_adam_ICLR15} and sample 8,192 rays per iteration. More details of settings can be found in our supplementary material.

\subsection{Comparisons}

\begin{table*}[t!]
\setlength{\tabcolsep}{4pt} %
\centering
\caption{\small{\textbf{Quantitative comparison.} We report LPIPS (lower is better) and PSNR/SSIM (higher is better) on eight dynamic scenes of the D-NeRF dataset.}}
\label{table:quant}
\resizebox{\textwidth}{!}{%
\begin{tabular}{lcccccccccccc}
\toprule
& \multicolumn{3}{c}{Hell Warrior} & \multicolumn{3}{c}{Mutant} & \multicolumn{3}{c}{Hook} & \multicolumn{3}{c}{Bouncing Balls} \\
Methods 
& PSNR$\uparrow$  & SSIM$\uparrow$ & LPIPS$\downarrow$
& PSNR$\uparrow$  & SSIM$\uparrow$ & LPIPS$\downarrow$
& PSNR$\uparrow$  & SSIM$\uparrow$ & LPIPS$\downarrow$
& PSNR$\uparrow$  & SSIM$\uparrow$ & LPIPS$\downarrow$\\
\cmidrule(lr){1-1}  \cmidrule(lr){2-4} \cmidrule(lr){5-7} \cmidrule(lr){8-10} \cmidrule(lr){11-13}
    D-NeRF\,(half$^1$)\cite{pumarola2021_dnerf_cvpr21}
    & 25.03 & \textBF{0.951} & \textBF{0.069}
    & 31.29 & 0.974 & 0.027 
    & 29.26 & \textBF{0.965} & 0.117
    & \textBF{38.93} & \textBF{0.990} & \textBF{0.103} \\
    NDVG\,(half$^1$) 
    & \textBF{25.53} & 0.949 & {0.073}
    & \textBF{35.53} & \textBF{0.988} & \textBF{0.014}
    & \textBF{29.80} & \textBF{0.965} & \textBF{0.037}
    & 34.58 & 0.972 & 0.114 \\
    \cmidrule(lr){1-1}  \cmidrule(lr){2-4} \cmidrule(lr){5-7} \cmidrule(lr){8-10} \cmidrule(lr){11-13}
    NDVG\,(w/o occ$^2$)
    & 25.16 & 0.956 & 0.067
    & 34.14 & \textBF{0.980} & 0.026
    & 29.88 & \textBF{0.963} & 0.047
    & 37.14 & 0.986 & 0.080 \\
    NDVG\,(w/o grid$^3$)
    & 26.45 & 0.959 & \textBF{0.065}
    & \textBF{34.42} & \textBF{0.980} & \textBF{0.025}
    & 29.08 & 0.956 & 0.050
    & \textBF{37.78} & \textBF{0.988} & 0.063 \\
    NDVG\,(w/o refine$^4$)
    & 24.30 & 0.946 & 0.092
    & 28.59 & 0.940 & 0.070
    & 26.85 & 0.935 & 0.081
    & 28.17 & 0.954 & 0.178 \\
    NDVG\,(w/o filter$^5$)
    & 19.55 & 0.927 & 0.104
    & 31.75 & 0.961 & 0.051
    & 27.71 & 0.948 & 0.068
    & 35.47 & \textBF{0.988} & \textBF{0.054} \\
    NDVG\,(full$^6$) 
    & \textBF{26.49} & \textBF{0.960} & {0.067}
    & 34.41 & \textBF{0.980} & 0.027 
    & \textBF{30.00} & \textBF{0.963} & \textBF{0.046}
    & 37.52 & 0.987 & 0.075 \\

\midrule
& \multicolumn{3}{c}{Lego} & \multicolumn{3}{c}{T-Rex} & \multicolumn{3}{c}{Stand Up} & \multicolumn{3}{c}{Jumping Jacks}  \\ 
Methods
& PSNR$\uparrow$  & SSIM$\uparrow$ & LPIPS$\downarrow$
& PSNR$\uparrow$  & SSIM$\uparrow$ & LPIPS$\downarrow$
& PSNR$\uparrow$  & SSIM$\uparrow$ & LPIPS$\downarrow$
& PSNR$\uparrow$  & SSIM$\uparrow$ & LPIPS$\downarrow$\\  
\cmidrule(lr){1-1}  \cmidrule(lr){2-4} \cmidrule(lr){5-7} \cmidrule(lr){8-10} \cmidrule(lr){11-13}
    D-NeRF\,(half) \cite{pumarola2021_dnerf_cvpr21}
    & 21.64 & 0.839 & 0.165
    & \textBF{31.76} & \textBF{0.977} & \textBF{0.040}
    & 32.80 & 0.982 & \textBF{0.021}
    & \textBF{32.80} & \textBF{0.981} & \textBF{0.037} \\
    NDVG\,(half)
    & \textBF{25.23} & \textBF{0.931} & \textBF{0.049}
    & 30.15 & 0.967 & 0.047
    & \textBF{34.05} & \textBF{0.983} & 0.022
    & 29.45 & 0.960 & 0.078 \\
    \cmidrule(lr){1-1}  \cmidrule(lr){2-4} \cmidrule(lr){5-7} \cmidrule(lr){8-10} \cmidrule(lr){11-13}
    NDVG\,(w/o occ)
    & 24.77 & 0.935 & 0.059
    & 32.57 & \textBF{0.979} & \textBF{0.031}
    & 18.78 & 0.899 & 0.112
    & 30.87 & 0.973 & 0.044 \\
    NDVG\,(w/o grid)
    & 24.18 & 0.916 & 0.078
    & 31.64 & 0.976 & 0.034
    & 32.99 & 0.980 & \textBF{0.027}
    & 30.64 & 0.971 & 0.044 \\
    NDVG\,(w/o refine)
    & 23.30 & 0.851 & 0.167
    & 27.35 & 0.940 & 0.079
    & 29.80 & 0.965 & 0.051
    & 26.13 & 0.920 & 0.150 \\
    NDVG\,(w/o filter)
    & 22.75 & 0.887 & 0.140
    & 28.58 & 0.952 & 0.067
    & 32.36 & 0.976 & 0.035
    & 28.19 & 0.957 & 0.077 \\
    NDVG\,(full)
    & \textBF{25.04} & \textBF{0.940} & \textBF{0.053}
    & \textBF{32.62} & 0.978 & 0.033
    & \textBF{33.22} & \textBF{0.979} & 0.030
    & \textBF{31.25} & \textBF{0.974} & \textBF{0.040} \\

\bottomrule %
\multicolumn{6}{l}{$^1$~\footnotesize{half: using half resolution of the original dataset images}} & \multicolumn{7}{l}{$^4$~\footnotesize{w/o refine: only using coarse training stage}} \\
\multicolumn{6}{l}{$^2$~\footnotesize{w/o occ: not using occlusion reasoning}} & \multicolumn{7}{l}{$^5$~\footnotesize{w/o filter: not using coarse training,direct optimize fine module}} \\
\multicolumn{6}{l}{$^3$~\footnotesize{w/o grid: not using deformation feature grid, only deform MLP}} & 
\multicolumn{7}{l}{$^6$~\footnotesize{full: using full resolution of the original dataset images}}
\end{tabular}
}

\end{table*}

\begin{table*}[t!]
\setlength{\tabcolsep}{4pt} %
\centering
\caption{\small{\textbf{Quantitative comparison on real scenes.}}}
\label{table:hyperquant3}
\resizebox{\textwidth}{!}{%
\begin{tabular}{lccccccccccc}
\toprule
&  & \multicolumn{2}{c}{3D Printer} & \multicolumn{2}{c}{Broom} & \multicolumn{2}{c}{Chicken} & \multicolumn{2}{c}{Peel Banana} & \multicolumn{2}{c}{Mean}\\
Methods
& Time
& PSNR$\uparrow$  & MS-SSIM$\uparrow$
& PSNR$\uparrow$  & MS-SSIM$\uparrow$
& PSNR$\uparrow$  & MS-SSIM$\uparrow$
& PSNR$\uparrow$  & MS-SSIM$\uparrow$
& PSNR$\uparrow$  & MS-SSIM$\uparrow$\\
\cmidrule(lr){1-1}  \cmidrule(lr){2-2} \cmidrule(lr){3-4} \cmidrule(lr){5-6} \cmidrule(lr){7-8} \cmidrule(lr){9-10}  \cmidrule(lr){11-12}
    NeRF~\cite{mildenhall2020_nerf_eccv20}
    & $\sim$ hours
    & 20.7 & 0.780 
    & 19.9 & 0.653 
    & 19.9 & 0.777 
    & 20.0 & 0.769 
    & 20.1 & 0.745 \\
    
    NV~\cite{Lombardi2019_NV_ACMTG}
    & $\sim$ hours
    & 16.2 & 0.665 
    & 17.7 & 0.623 
    & 17.6 & 0.615 
    & 15.9 & 0.380 
    & 16.9 & 0.571 \\
    
    NSFF~\cite{li2021_nsff_cvpr21}
    & $\sim$ hours
    & 27.7 & 0.947
    & 26.1 & 0.871 
    & 26.9 & 0.944
    & 24.6 & 0.902  
    & 26.3 & 0.916 \\
    
    Nerfies~\cite{park2021_nerfies_iccv21}
    & $\sim$ hours
    & 20.6 & 0.830 
    & 19.2 & 0.567 
    & 26.7 & 0.943 
    & 22.4 & 0.872 
    & 22.2 & 0.803 \\
    
    HyperNeRF~\cite{park2021hypernerf}
    & $\sim$ hours
    & 20.0 & 0.821 
    & 19.3 & 0.591 
    & 26.9 & 0.948 
    & 23.3 & 0.896 
    & 22.4 & 0.814 \\
    
    TiNeuVox~\cite{fang2022_TANV_arxiv}
    & 30 mins
    & 22.8 & 0.841 
    & 21.5 & 0.686 
    & 28.3 & 0.947 
    & 24.4 & 0.873  
    & 24.3 & 0.837 \\

\midrule

    NDVG~(Ours)
    & 35 mins
    & 22.4 & 0.839 
    & 21.5 & 0.703 
    & 27.1 & 0.939 
    & 22.8 & 0.828  
    & 23.3 & 0.823 \\

\bottomrule %

\end{tabular}
}

\end{table*}

\paragraph{Quantitative evaluation on the dataset} 
We first quantitatively compare the results in~\Tref{table:quant}. To compare with D-NeRF\cite{pumarola2021_dnerf_cvpr21}, we set the same 400$\times$400 image resolution, and present average results on each scene assessed by metrics which are mentioned above. According to \Tref{table:quant}, we could see that our method NDVG achieves comparable results with D-NeRF~\cite{pumarola2021_dnerf_cvpr21} for all three metrics. For real scenes, we test on four scenes, namely the \emph{Peel Banana}, \emph{Chicken}, \emph{Broom} and \emph{3D Printer} following HyperNeRF \cite{park2021hypernerf}. We follow the same settings of the experiments of TiNueVox \cite{fang2022_TANV_arxiv} and report the metrics of PSNR and MS-SSIM in \Tref{table:hyperquant3} (results of other methods are taken from the TiNuxVox paper). 
As shown in \Tref{table:hyperquant3}, our method could achieve comparable or even better results and at least 10x faster, which clearly demonstrates the effectiveness of our method.

\begin{table*}[!tbp]
    \centering
    \caption{{\bf Training time and rendering speed comparison.}
    We report these using the public code of D-NeRF~\cite{pumarola2021_dnerf_cvpr21} on the same device (RTX 3090 GPU) with our method. We include the mean PSNR across eight scenes in the D-NeRF dataset~\cite{pumarola2021_dnerf_cvpr21} for comparison of synthesized image quality. Our method could achieve good PSNR, while spend much less optimization time and have faster rendering speed.}
    \label{table:speed}
    \resizebox{0.7\textwidth}{!}{%

    \bgroup
    \small
    
    \begin{tabular}{lccc}
    \toprule %
    Methods & PSNR$\uparrow$ & Training Time (s/scene)$\downarrow$ & Rendering Speed (s/img)$\downarrow$ \\
    \cmidrule(lr){1-1} \cmidrule(lr){2-2} \cmidrule(lr){3-3} \cmidrule(lr){4-4} 
    NeRF(half)$^{\dagger}$~\cite{mildenhall2020_nerf_eccv20} & 19.00 & 60185 & 4.5 \\
    D-NeRF(half)~\cite{pumarola2021_dnerf_cvpr21} & 30.02 & 99034 & 8.7 \\
    NDVG(half) & \textBF{30.32} & \textBF{1380} & \textBF{0.4} \\
    \cmidrule(lr){1-1} \cmidrule(lr){2-2} \cmidrule(lr){3-3} \cmidrule(lr){4-4}
    NDVG(w/o refine) & 26.73 & \textBF{708} & 2.6  \\
    NDVG(w/o filter) & 27.85 & 2487 & 3.5  \\
    NDVG(full) & \textBF{31.08} & 2087 & \textBF{1.7} \\
    \bottomrule %
    \multicolumn{4}{l}{$^{\dagger}$~\footnotesize{We use implementation of D-NeRF\cite{pumarola2021_dnerf_cvpr21} to train NeRF on dynamic dataset}}
    \end{tabular}
    \egroup
    }

\end{table*}

\begin{figure}[t] \centering
    \includegraphics[width=0.95\textwidth]{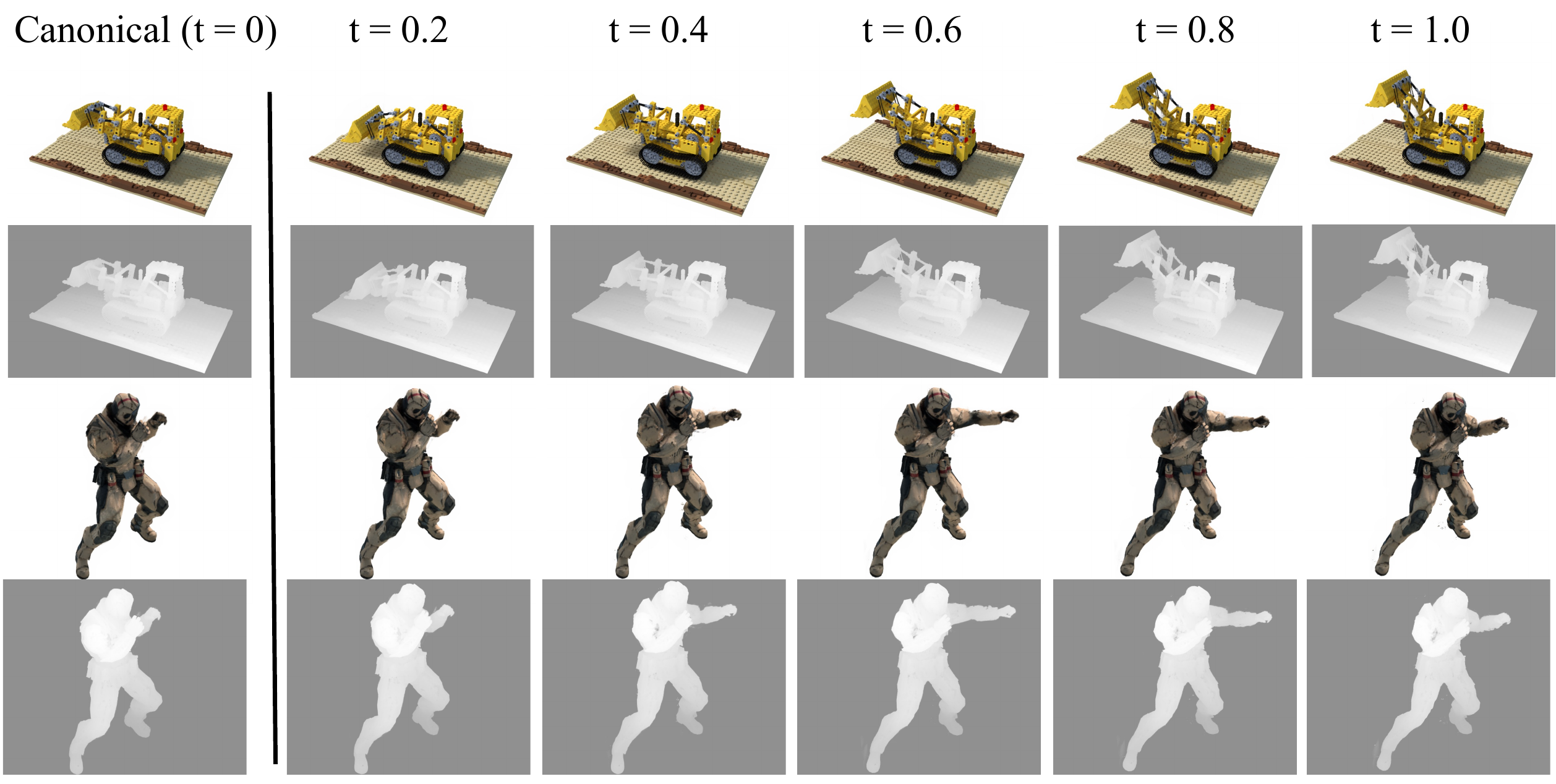}
    \caption{\textbf{Learned Geometry.} We show examples of geometries learned by our model. For each, we show rendered images and corresponding disparity under two novel views and six time steps.} \label{fig:visgeo}
\end{figure}

\begin{figure}[t] \centering
    \includegraphics[width=0.95\textwidth]{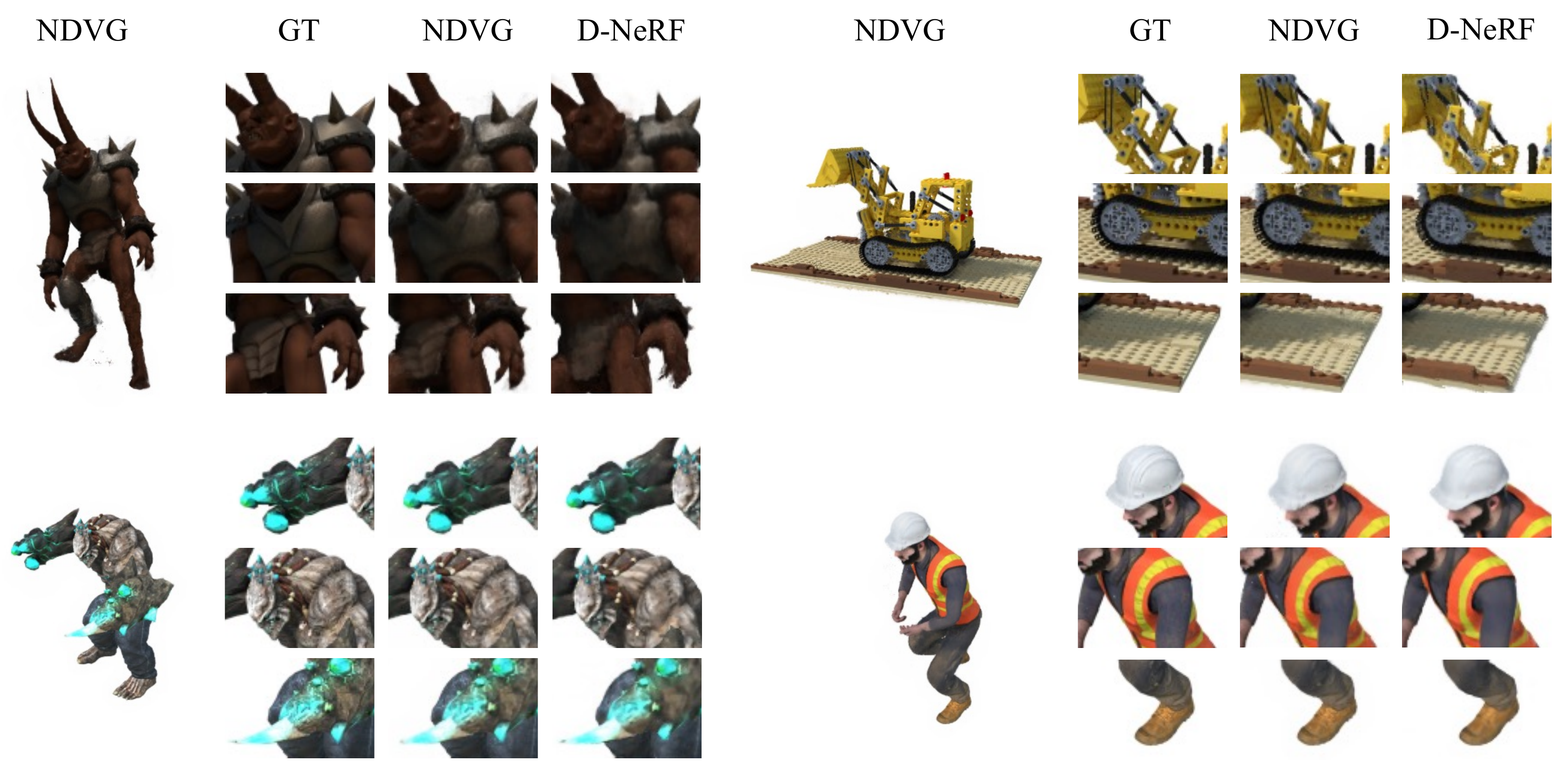}
    \caption{\textbf{Qualitative Comparison.} Synthesized images on test set of the dataset. For each scene, we show an image rendered at novel view, and followed by zoom in of ground truth, our NDVG, and D-NeRF~\cite{pumarola2021_dnerf_cvpr21}.} \label{fig:compare}
\end{figure}

\paragraph{Training time and rendering speed comparison}
The key contribution of our work is to accelerate the optimization speed of novel view synthesis models on dynamic scenes. In~\Tref{table:speed}, our method with the same half resolution setting with D-NeRF\cite{pumarola2021_dnerf_cvpr21}, achieves 70$\times$ faster convergence with an even higher average PSNR. Though our main purpose is to speed up training, the proposed method also has a reasonably fast rendering speed, compared to complete neural network based model. According to ~\Tref{table:speed}, our method has a 20$\times$ faster rendering speed compared with D-NeRF\cite{pumarola2021_dnerf_cvpr21}.

For real scenes in \Tref{table:hyperquant3}, compared with previous methods without acceleration which takes hours or even days to train, our method could finish training in 35 minutes which is at least 10x faster. Compared with the concurrent research TiNueVox \cite{fang2022_TANV_arxiv}, which also aims to speed up training, we could achieve comparable results with the same training time, without using cuda acceleration for ray points sampling and total variation computation.

\paragraph{Qualitative comparison}
We provide some visualization of the learned scene representation in~\fref{fig:visgeo}. 
We can see that our method can successfully recover the canonical geometry, and render high-quality dynamic sequences. The results indicate that our method can faithfully model the motion of the dynamic scenes.

In~\fref{fig:compare}, we show the rendering results of more difficult situations and compare them with the results of D-NeRF~\cite{pumarola2021_dnerf_cvpr21}. 
Our method achieves comparable or even better image results using only $1/70$ of the training time. If zoom in for more details, we could see that our method actually recovers more high-frequency details, taking the armour and the cloth of the worker as examples.

\subsection{Method Analysis}
In this section, we aim to study and prove the effectiveness of three designs in our proposed method: the occlusion reasoning, the coarse-to-fine optimization strategy, and the deformation feature grid.

\begin{figure}[t] \centering
    \includegraphics[width=\textwidth]{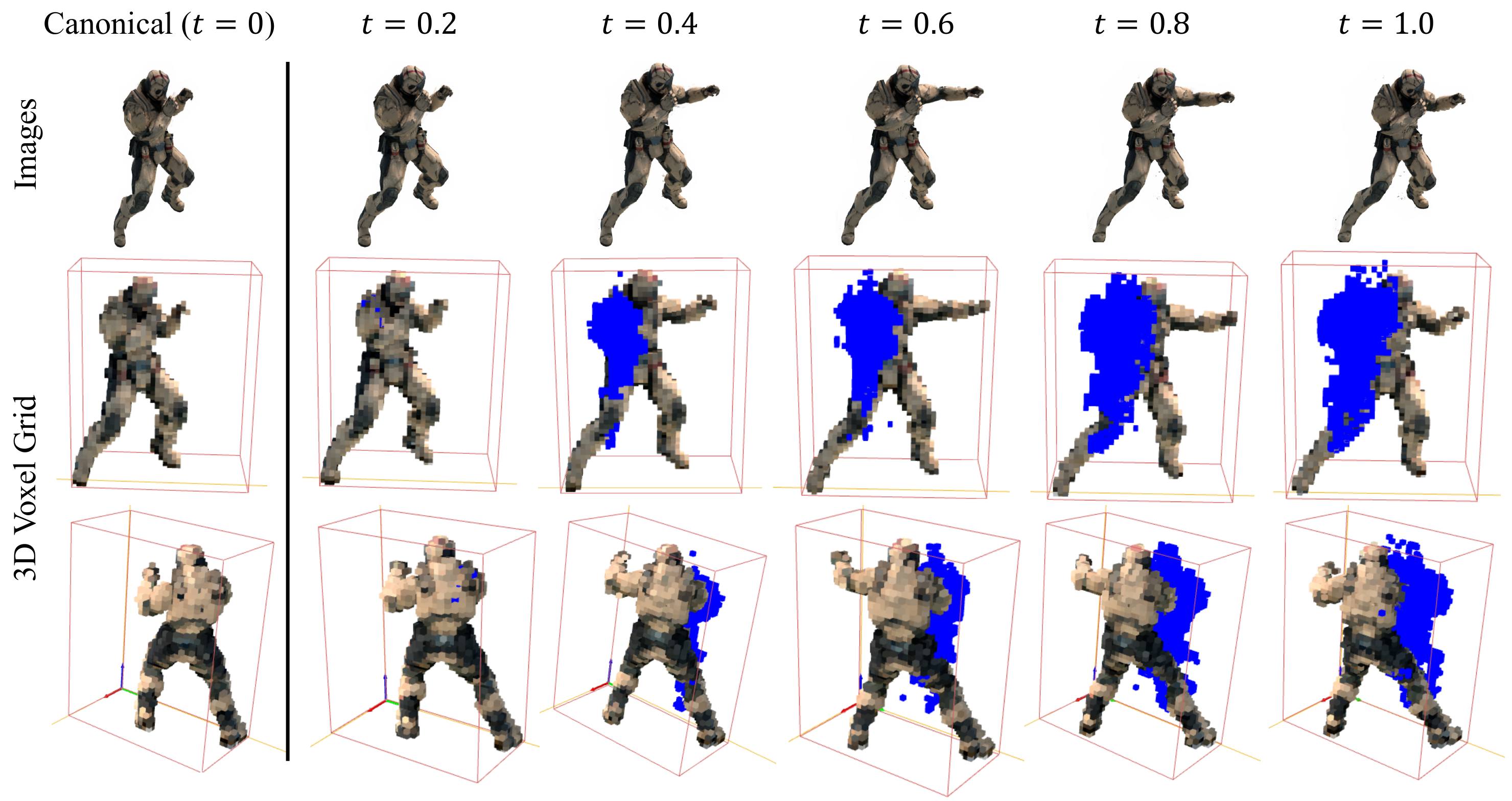}
    \caption{\textbf{Occlusion Estimation Visualization.} We visualize the estimated occlusion points at different times in blue color. The first row shows the images at each time step which gives an insight of the motion. We warp the canonical grids into corresponding time step by the deformation estimated by \deformmodule. We visualize points with their rgb colors whose density is over 0.8.} \label{fig:occlusion}
\end{figure}

\paragraph{Occlusion Reasoning}
We show that occlusion reasoning is critical in~\sref{sec:occlusion} for the canonical-based pipeline under the assumption that the empty point is static.
Results of NDVG~(full) and NDVG~(w/o occ) in~\Tref{table:quant} compare the quantitative results of models with and without occlusion reasoning. 
We can see that the model with occlusion reasoning achieves more accurate results, verifying the effectiveness of our method.

We also visualize the estimated occlusion at different time steps in~\fref{fig:occlusion}. 
We warp the canonical grid (density grid and color grid) into the different time steps, and show the points with corresponding colors if they are object points. %
And we show the occluded points estimated by \deformmodule in blue. 
In~\fref{fig:occlusion}, while the character pushes out the punch and the body moves forward, the empty points behind his back are estimated as being occluded which fits the actual situation.
These results indicate the estimated occlusion is accurate.

\paragraph{Effectiveness of Coarse-to-fine Optimization}
The coarse-to-fine optimization could not only improve the image quality rendered by our model, but also help speed up the training and rendering process significantly. To prove the efficiency of the coarse-to-fine optimization, we conduct two extra experiments (see~\Tref{table:quant}). 
The first one is NDVG~(w/o refine), which only contains the coarse training stage. 
As NDVG~(w/o refine) has limited grid resolution and does not have view-independent color representation, the performance is expected to be low.
The second one is NDVG~(w/o filter), which is not initialized by coarse training and directly begins from scratch for fine stage. 
As NDVG~(w/o filter) does not have a trained coarse model for object region location, the fine model could not shrink the bounding box of the scene which means most of the grid space is wasted, which lead to obvious worse results. Also, the empty points cannot be filtered, which could increase training time and decrease rendering speed significantly, which is evidenced in \Tref{table:speed}.

\paragraph{Deformation Feature Gird}
In our \deformmodule, we use a deformation feature grid to encode dynamic information of 3D points, which will be decoded by a light-weight MLP to regress the deformation. 
A light-weight MLP is sufficient to model the complex motion of scenes by integrating with this feature grid.
\Tref{table:quant} shows that the method integrated with a deformation feature grid, NDVG~(full), achieves better novel-view synthesis results than the one without, NDVG~(w/o grid). This result clearly demonstrates the effectiveness of the proposed deformation feature grid. Please refer to our supplementary material for further study of the deformation feature grid design.

\section{Conclusions}

In this paper, we have presented a fast optimization method for dynamic view synthesis based on a hybrid implicit-explicit representation.
Our method consists of a \deformmodule to map a 3D point in observation space to canonical space, and a \canonicalmodule to represent the scene geometry and appearance. In each module, explicit dense grids and the light-weight MLP are used for fast feature interpolation and decoding, which significantly accelerates the optimization time compared with methods that rely on heavy MLP queries. Moreover, occlusion is explicitly modeled to improve the rendering quality. Experiments show that our method only requires $30$ minutes to converge on a dynamic scene. Compared with the existing D-NeRF method, our method achieves a $70\times$ acceleration with comparable rendering quality.

\paragraph{Limitation} Despite our method greatly speeds up the training of radiance field methods for dynamic view synthesis, it is mainly designed for bounded scenes. In the future, we will extend our method to deal with real-world unbounded scenes (\eg, unbounded 360 and face-forward scenes).

\sloppy
\subsubsection{Acknowledgements}
This work was supported in part by the National Natural Science Foundation of China (Nos. 61871325, 61901387, 62001394, 62202409) and the National Key Research and Development Program of China (No. 2018AAA0102803).

\bibliographystyle{splncs04}
\bibliography{9_References}

\end{document}


\title{Supplementary Material for \\ ``Neural Deformable Voxel Grid for \\ Fast Optimization of Dynamic View Synthesis''}
\titlerunning{NDVG}
\author{Xiang Guo\inst{1*} \and
Guanying Chen\inst{2*} \and
Yuchao Dai\inst{1}$^{\dag}$ \and
Xiaoqing Ye\inst{3} \and \\
Jiadai Sun\inst{1} \and
Xiao Tan\inst{3} \and
Errui Ding\inst{3}
}
\authorrunning{X. Guo et al.}
\institute{
Northwestern Polytechnical University \and
FNii and SSE, CUHK-Shenzhen \and
Baidu Inc. \\
\email{\{guoxiang,sunjiadai\}@mail.nwpu.edu.cn},
\email{chenguanying@cuhk.edu.cn},
\email{daiyuchao@nwpu.edu.cn}, 
\email{\{yexiaoqing,dingerrui\}@baidu.com},
\email{tanxchong@gmail.com}\\
}
\maketitle              %

\blfootnote{%
* Authors contributed equally to this work. $^{\dag}$ Yuchao Dai is the corresponding author.}

To the best of our knowledge, we are one of the first to speed up the training of dynamic NeRF by integrating the voxel-grid optimization with a deformable radiance field. 
Note that making this idea work is non-trivial.
One hand, it is difficult to optimize the voxel gird to achieve reasonable results due to its discontinuity nature, which requires us to design different constraints while keeping the model capacity. 
On the other hand, to further speed up the training, we design a coarse-to-fine strategy with filtering strategies specifically designed for dynamic scenes. 
We believe our method, which can speed up training by 70$\times$ while maintaining comparable rendering quality, is a good step forward in the fast optimization of dynamic view synthesis. Here, we provide more details and analysis of the design, followed with more experiment results, to further discuss the method we proposed.
\section{More Details for the Proposed Method}

\subsection{Empty space filtering}

The following two strategies are used to speed up the training. 
First, we locate the smallest box region that fully covers the whole scene, which allows us to make full use of the grid resolution. 
To achieve this, we query all grid vertices of the coarse deformation feature grid in all training time steps to get the alpha values. 
We assume a point belongs to an object point rather than the empty space if its alpha value is larger than a predefined threshold. 
Using alpha values of all grid vertices at all training time steps, we could find the smallest bounding box for the dynamic scene. 

Second, we filter out the sampled points in a ray if they are recognized as empty points, reducing the query numbers to the light-weight MLP.
For sampled points in the \deformmodule, as we assume empty points to be static, we filter out sampled points whose alpha is always below a threshold during all training times. 
For sampled points in the \canonicalmodule, we only filter out the empty points at the canonical time.

\subsection{Fine Model Design}

The MLP architecture of \deformmodule in the fine model is the same as in the coarse model, such that the trained network weights in the coarse \deformmodule can be used for initializing the fine model.
We initialize the deformation feature grid with the optimized feature grid in the coarse model by interpolation. 
This initialization can result in higher performance and shorten the training time.

We model the view-dependent effect in the fine model by using a light-weight MLP to decode the interpolated color feature. We also use a progressive scale training, which doubles the voxel resolution after some training iterations~\cite{sun2021direct}.

\subsection{Network Architecture}

\begin{figure}[h] \centering
    \includegraphics[width=0.8\textwidth]{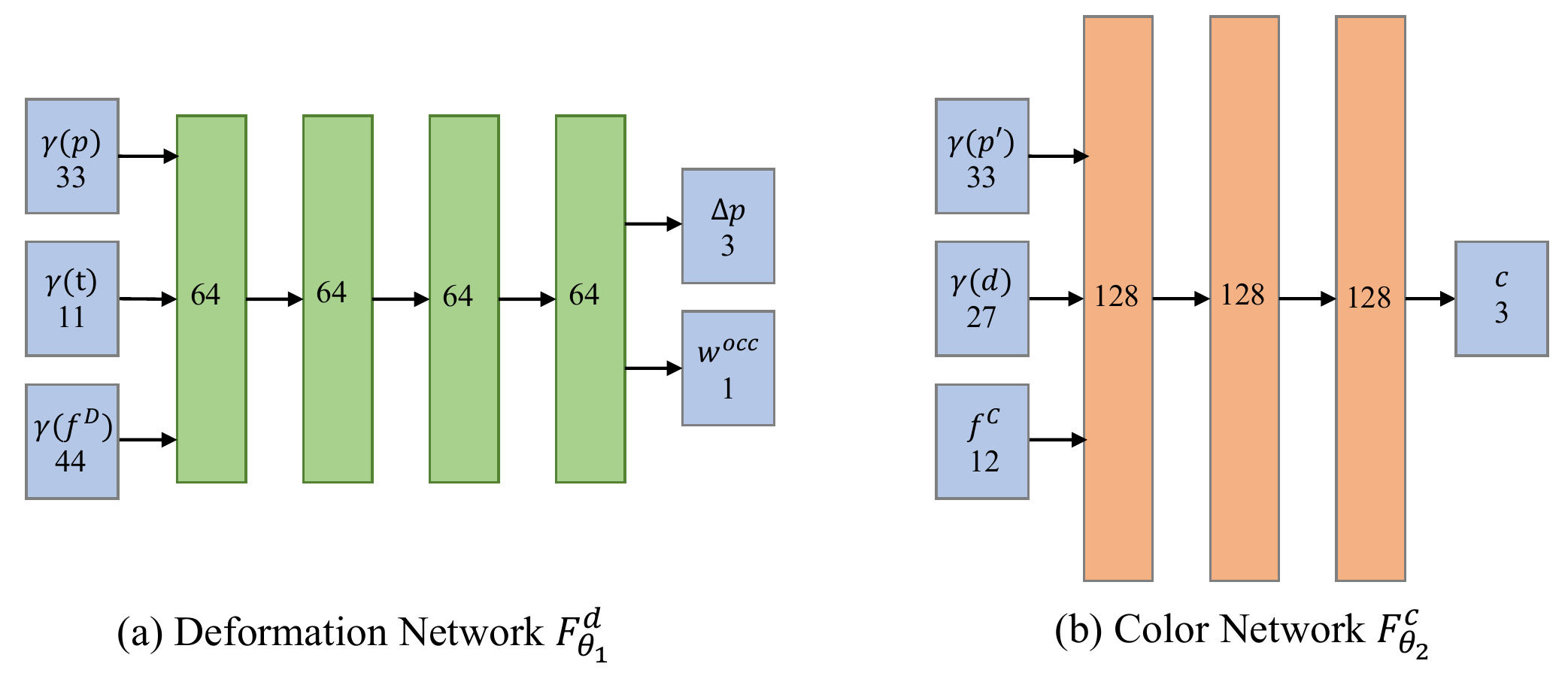}
    \caption{\textbf{Network Architecture.} We show architectures of our light-weight deformation network and color network.} \label{fig:suppnetarch}
\end{figure}

We show the architecture of our networks in \fref{fig:suppnetarch}. 
For deformation network $\DNet$, we use four layers of fully connected layer with the width set to 64. 
For the input layer, the dimensions of position, time and feature are 33, 11 and 44. 

For the color network $\CNet$, we use three layers of fully connected layer with the width set to 128. 
For the input layer, the dimensions of position, view direction, and feature are 33, 27 and 12. Between the fully connected layer, we use ReLU as activation functions. For the occlusion output $w^{occ}$ and color output $c$, we apply sigmoid activation to transfer outputs into range of $(0, 1)$.

\subsection{Hyper-parameter Settings in Experiments}
As described in Eq.~(16) of the paper, the overall loss function for the coarse stage and the fine stage can be written as
\begin{gather}
\loss{} =\loss{photo}+\weight{ptc}{}\cdot \loss{ptc} + \weight{bg}{} \cdot \loss{bg} + \weight{d\_norm}{} \cdot \loss{d\_norm} + \weight{d\_tv}{}\cdot \loss{d\_tv},
\end{gather}
where $\weight{ptc}{}$, $\weight{bg}{}$, $\weight{df\_norm}{}$, and $\weight{df\_tv}{}$ are weights to balance each component.

In the coarse stage, we empirically set $\weight{ptc}{}$, $\weight{bg}{}$, $\weight{df\_norm}{}$, and $\weight{df\_tv}{}$ to 0.1, 0.01, 0.1, and 1, respectively.
In the fine stage, these four weights were set to be smaller values as 0.01, 0.001, 0.01, and 1, respectively.
As the motions of `\texttt{Hell Warrior}', `\texttt{Jumping Jacks}' and `\texttt{T-Rex}' are larger, we used a smaller weight for the deformation norm regularization, with $\weight{df\_norm}{}$ equals to $0.01$ and $0.001$ for the coarse and fine stage.

\begin{figure}[!t] \centering
    \includegraphics[width=0.88\textwidth]{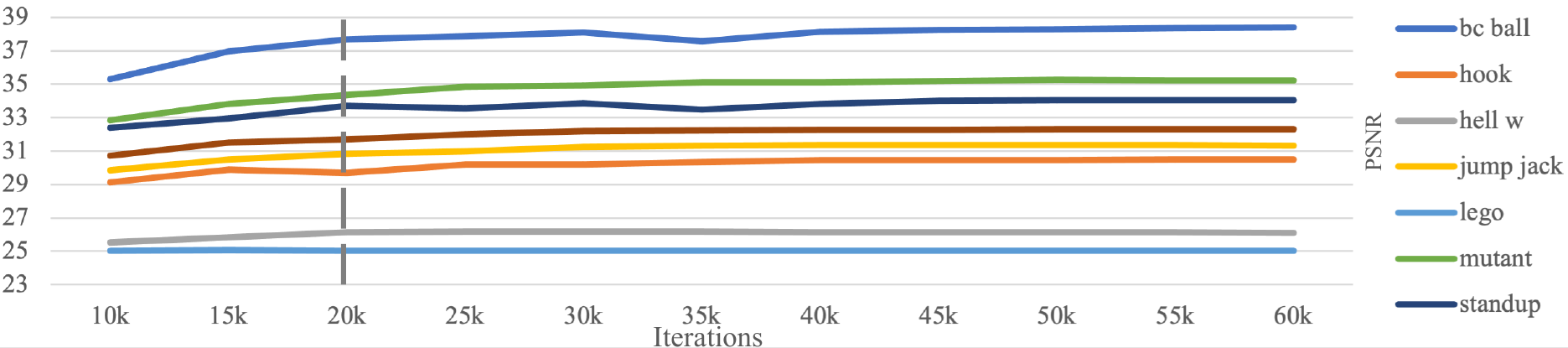}
    \caption{\footnotesize{PSNR with more training iterations}} \label{fig:moreite}
\end{figure}

As shown in \fref{fig:moreite}, the PSNR generally increase slightly after 20k iteratins, but will takes significantly more iterations to reach the max. To balance the PSNR and time consumption, we set the training iteration to 20k. 

\subsection{Discussion of Tips to Speed Up Training}

The purpose to use coarse-to-fine training is that we could get a relative good geometry and motion model quickly and other strategy like empty-space filtering is built on this coarse model to save training time in fine stage. In~\Tref{tab:moreablations}, $\text{NDGV}_{\text{(w/o filter)}}$
does not use any speed-up strategies and directly optimizes fine model without filtering. Under this setting, training speed, render speed and PSNR are all clearly worse than our full model $\text{NDGV}_{\text{(full)}}$.
On the other hand, the train speed of $\text{NDGV}_{\text{(w/o filter)}}$
is on the same level with $\text{NDGV}_{\text{(full)}}$, which means these tips are not the main factors to speed up training.

\begin{table}[!t]
\centering
\caption{More results of ablations \footnotesize{(\bf{half:400x400, full:800x800})}}
\label{tab:moreablations}
        \resizebox{0.78\textwidth}{!}{
    \large
    \begin{tabular}{*{6}{c}}
        \toprule
       Metrics & DNeRF &  NDGV & NDGV &  NDGV & NDGV \\
               & (half) &  (half) & (full, w/o filter) &  (full, w/o grid) & (full) \\
        \midrule
        train speed (s/scene)  & 99034 &  1380 &  2487 &  1450 &  2087 \\
        render speed (s/image) &   8.7 &   0.4 &   3.5 &   1.3 &   1.7 \\
        model size (M)         &  13.2 & 995.8 & 988.3 & 668.1 & 994.2 \\
        PSNR                   & 30.02 & 30.32 & 27.85 & 30.62 & 31.08 \\
        \bottomrule
        \multicolumn{3}{l}{w/o filter: without any filter methods} & \multicolumn{3}{l}{w/o grid: without deformation grid} \\
    \end{tabular}
    }

    \vspace{-0.5em}
\end{table}

\subsection{Illustration of Empty Space Filtering}

\begin{figure}[h] \centering
    \includegraphics[width=\textwidth]{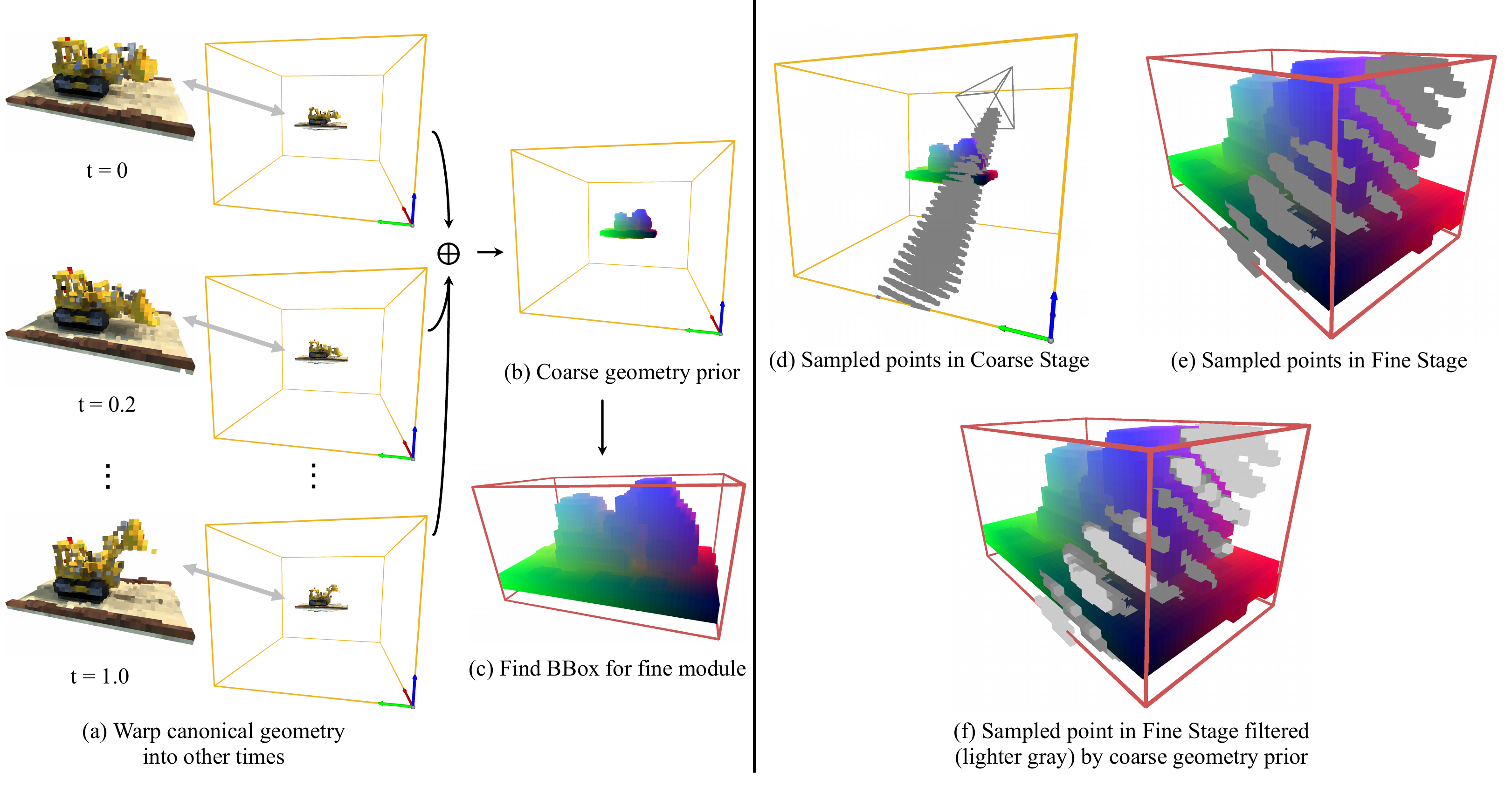}
    \caption{\textbf{Illustration of Empty Space Filtering.} (a) warps the learnt canonical geometry during coarse training stage into other times. Be aware of the moving part in red circle. (b) combines object points in all times to form a coarse geometry prior, indicating all possible positions that could be an object point along all times. (c) with coarse geometry prior, find a smallest bounding box which could cover all possible object points. The volume of the grid is significantly reduced. (d) shows a example of sampled point in Coarse Stage masked by coarse bounding box. (e) shows a example of sampled point in Fine Stage masked by fine bounding box. (f) shows sampled points in (e) but filtered by coarse geometry prior. Be aware that the sampled points in empty space are marked with lighter gray color.} \label{fig:suppfilter}%
\end{figure}

As described in Section 4.2 of the paper, we speed up the training with the empty space filtering strategy.
Here, we visualize our empty space filtering strategy in \fref{fig:suppfilter}. 
We first locate the smallest box region that fully covers the whole scene, which is denoted as coarse geometry prior (see \fref{fig:suppfilter}~(a)-(c)).
We obtain the coarse geometry prior by warping object points in a canonical space into all other times, so that we can identify all possible positions that could be object points in the world coordinates.
With this coarse geometry prior, we could reduce the volume of the grid in fine stage to avoid as many empty point as possible.

Second, we filter out the sampled points in a ray if they are recognized as empty points, reducing the query numbers to the light-weight MLP (see \fref{fig:suppfilter}~(d)-(f)). 
Specifically, we filter our sampled points whose alpha values are always below a threshold during all training times. 
All these operations significantly reduce the time consumption of training and rendering.

\section{More Analysis for the Proposed Method}

\subsection{Training, Rendering Speed and Model Size}

We show the training and render speed of different settings in~\Tref{tab:moreablations}.
To speed up training, we need to use a very small size MLP, which could harm the image quality. This is why we use a deformation feature grid to embed more information without adding too much complexity to the system. We can see that adding a deformation feature grid increases the PSNR at the cost of slightly more training time and bigger model size, which we believe is reasonable to balance these metrics. 

We report model size in~\Tref{tab:moreablations}. The model size of DNeRF is the smallest thanks to the compact representation of MLP, but it doesn't mean it is faster and cheaper to train. In fact, we could train faster on the same device compared with DNeRF and set a much bigger batch size with a resolution of around $150^3$ in our experiments.

In addition, We further show the training speed of each submodule in our system in \Tref{table:submtime}. According to \Tref{table:submtime}, we can notice a dramatic increase of time consumption of query deformation feature, deformation network forward and gradient computation process without our empty space filtering mechanism, since more empty points are involved in the optimization process without filtering.

\begin{table}[h]
    \centering
    \caption{{\bf Detailed Training Time of Each Module}.
    We report the training time in details of each module in \textbf{second/iteration}. \textbf{Query Feat} is query the deformation feature grid. \textbf{Deform Net} means deformation network forward. \textbf{Canonical Module} means the whole process in canonical module. \textbf{Render} means the render process after get the densities and colors of all sampled points. \textbf{Loss} means the computation of loss. \textbf{Backward} means the computation of the gradients. \textbf{Optimization} is the process of optimizing all parameters.}
    \label{table:submtime}
        \resizebox{0.98\textwidth}{!}{%

    \bgroup
    \small
    
    \begin{tabular}{lccccccc}
    \toprule %
    Methods & \multicolumn{2}{c}{Deformation Module} & Canonical Module & Render & Loss & Backward & Optimization \\
    \cmidrule(lr){2-3}
     & Query Feat & Deform Net & & & & & \\
    \cmidrule(lr){1-1} \cmidrule(lr){2-2} \cmidrule(lr){3-3} \cmidrule(lr){4-4} \cmidrule(lr){5-5} \cmidrule(lr){6-6} \cmidrule(lr){7-7} \cmidrule(lr){8-8}
    NDVG~(full) &
    9.23e-4 & 5.06e-3 & 5.28e-3 & 1.02e-3 & 5.61e-3 & 3.50e-2 & 7.23e-3 \\
    NDVG~(w/o filter) &
    3.28e-3 & 2.82e-2 & 4.93e-3 & 1.08e-3 & 5.36e-3 & 6.68e-2 & 7.36e-3 \\
    \bottomrule
    \end{tabular}
    \egroup
    }

\end{table}

\vspace{1cm}
\subsection{Ablation Study of Losses}

\begin{table*}[h]
    \centering
    \caption{{\bf Ablation Study of Losses}.
    }
    \label{table:losses}
        \resizebox{0.45\textwidth}{!}{%

    \bgroup
    \small
    
    \begin{tabular}{lccc}
    \toprule %
    Methods & PSNR$\uparrow$  & SSIM$\uparrow$ & LPIPS$\downarrow$ \\
    \cmidrule(lr){1-1} \cmidrule(lr){2-2} \cmidrule(lr){3-3} \cmidrule(lr){4-4}
    NDVG~(w/o $\loss{d\_tv}$) &
    30.45 & 0.963 & 0.061 \\
    NDVG~(w/o $\loss{d\_norm}$) &
    31.54 & 0.971 & 0.039 \\
    NDVG~(full) &
    31.08 & 0.970 & 0.039 \\
    \bottomrule
    \end{tabular}
    \egroup
    }

\end{table*}
\begin{figure}[h] \centering
    \includegraphics[width=\textwidth]{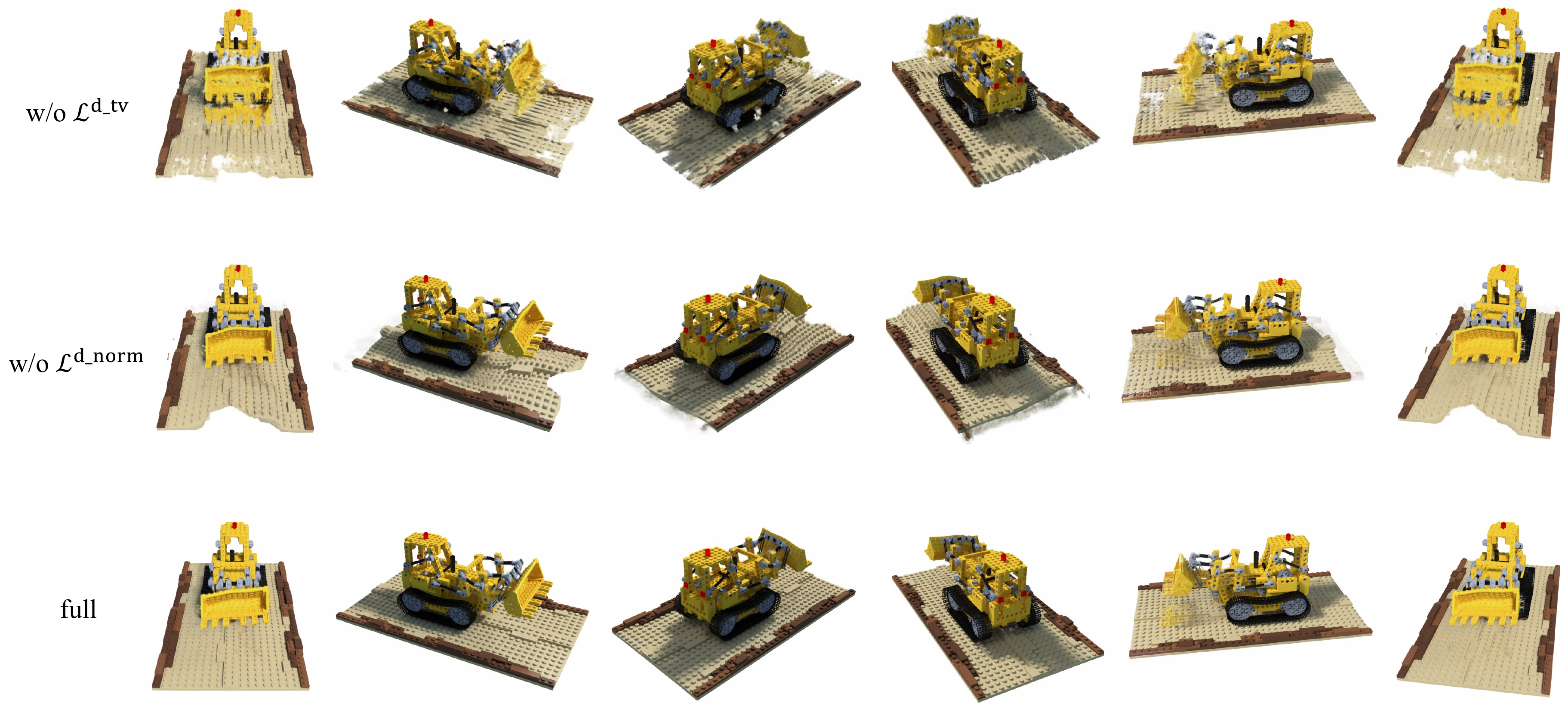}
    \caption{\textbf{The Reconstructed Canonical space with different Losses}.} \label{fig:supplosses}
\end{figure}

We establish the ablation study of the losses we use during training. We show the results of models trained without the total variation loss $\loss{d\_tv}$ and without the deformation normalization loss $\loss{d\_norm}$ in \Tref{table:losses}. According to  \Tref{table:losses}, there is a significant drop without $\loss{d\_tv}$. It is worth to mention that, even it achieve slightly better results without deformation normalization loss $\loss{d\_norm}$, the geometry of canonical module degenerated due to lack of normalization of the canonical space, which is show in \fref{fig:supplosses}. As shown in \fref{fig:supplosses}, without $\loss{d\_tv}$, the canonical images are shattered without this spatial smooth term. Also, without $\loss{d\_norm}$, the canonical images are distorted. Since there are no constraints on deformation estimation, there is more freedom of the deformation between spaces at other times and the canonical space, which could cause distortions in the canonical space.

\subsection{Deformation Grid Resolution}

\begin{figure}[ht] \centering
    \includegraphics[width=0.5\textwidth]{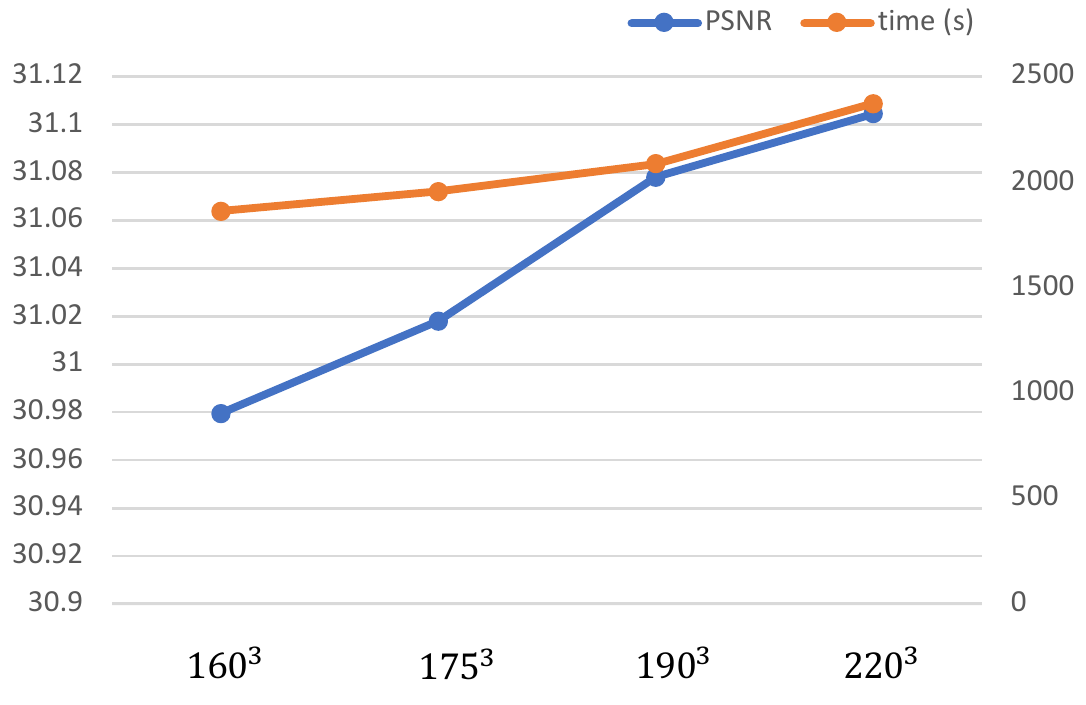}
    \caption{\textbf{Effects of Using Different Resolutions for the Deformation Grid}.} \label{fig:suppresolution}
\end{figure}

We test the different deformation grid resolution (total voxel number) and analysis the corresponding images quality and training time consumption, which is show in \fref{fig:suppresolution}. With the total number of voxels is increasing, the overall trends of PSNR and training time consumption are also increase. We choose the final number of voxels as $190^3$ to balance the image quality and training time consumption.

\subsection{Visualization of the Feature Grid}

\begin{figure}[t] \centering
    \includegraphics[scale=0.5]{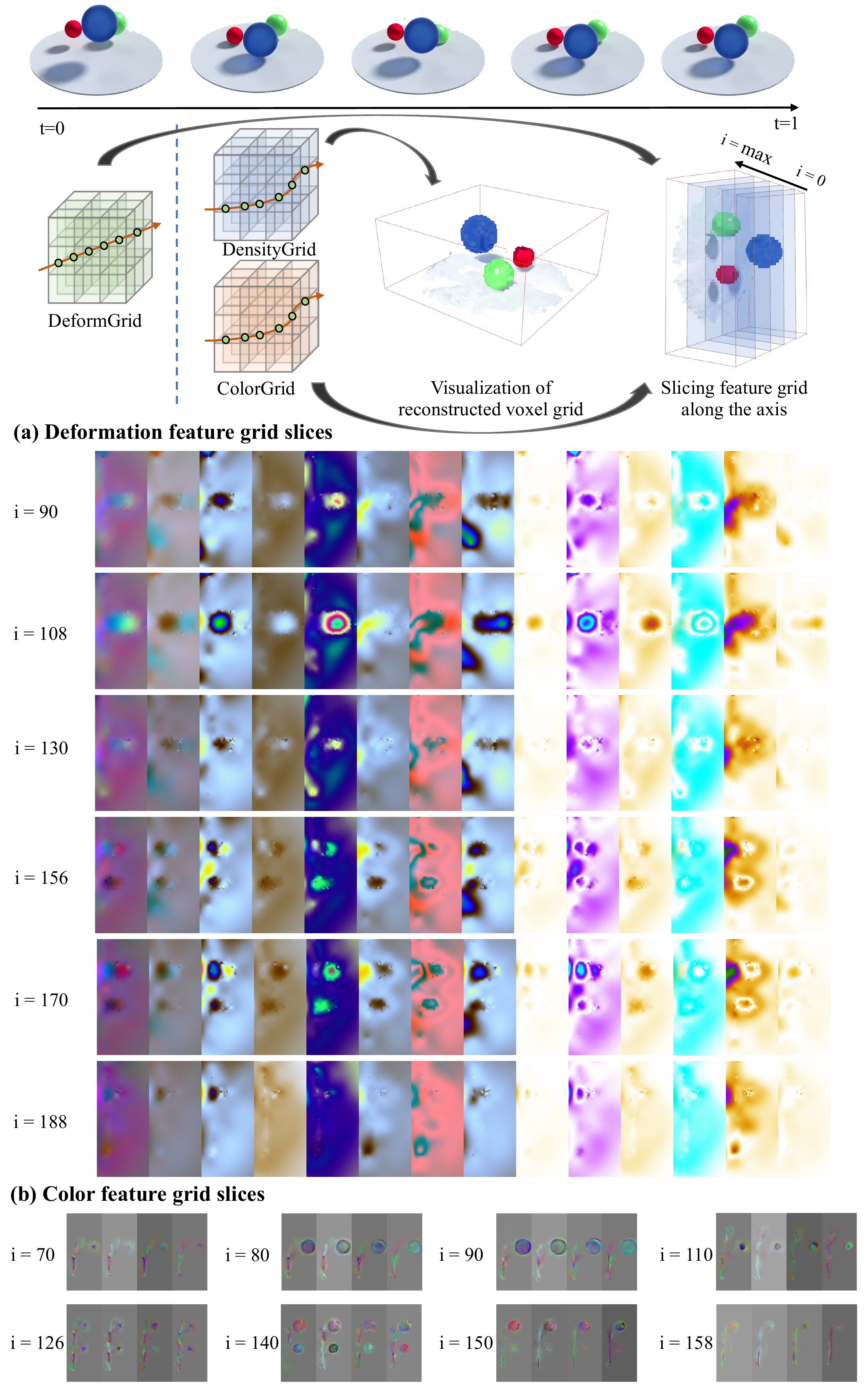}
    \caption{\textbf{Visualization of Feature Grid.} We slice the feature grid along the axis to get feature slices. Then we visualize the feature slices every three dimension as rgb images after normalization. Best viewed in color and zoom in for details.} \label{fig:suppvizfeature}
\end{figure}

The feature grids, including deformation feature grid in Deformation Module and color feature grid in Canonical Module, play important roles in our pipeline. We show the ablation study in paper with quantitative results. Here we visualize the feature learnt from training in \fref{fig:suppvizfeature}. As shown of the first row in \fref{fig:suppvizfeature}, we slice the feature grid along the axis $x$. We then treat the sliced feature slice as stacked rgb images and visualize them after normalization. In terms of deformation of feature grid slices, some of the columns (like the 4th and 5th) show clean clues of objects. Other columns (like the 6th and 7th), show clean clues of deformations of the objects (bouncing trajectory). In terms of color feature grid slices, the visualizations show clean clues of objects and no clues of movements, because it is defined in canonical space.

\section{More Results}
\begin{figure}[t] \centering
    \includegraphics[width=0.9\textwidth]{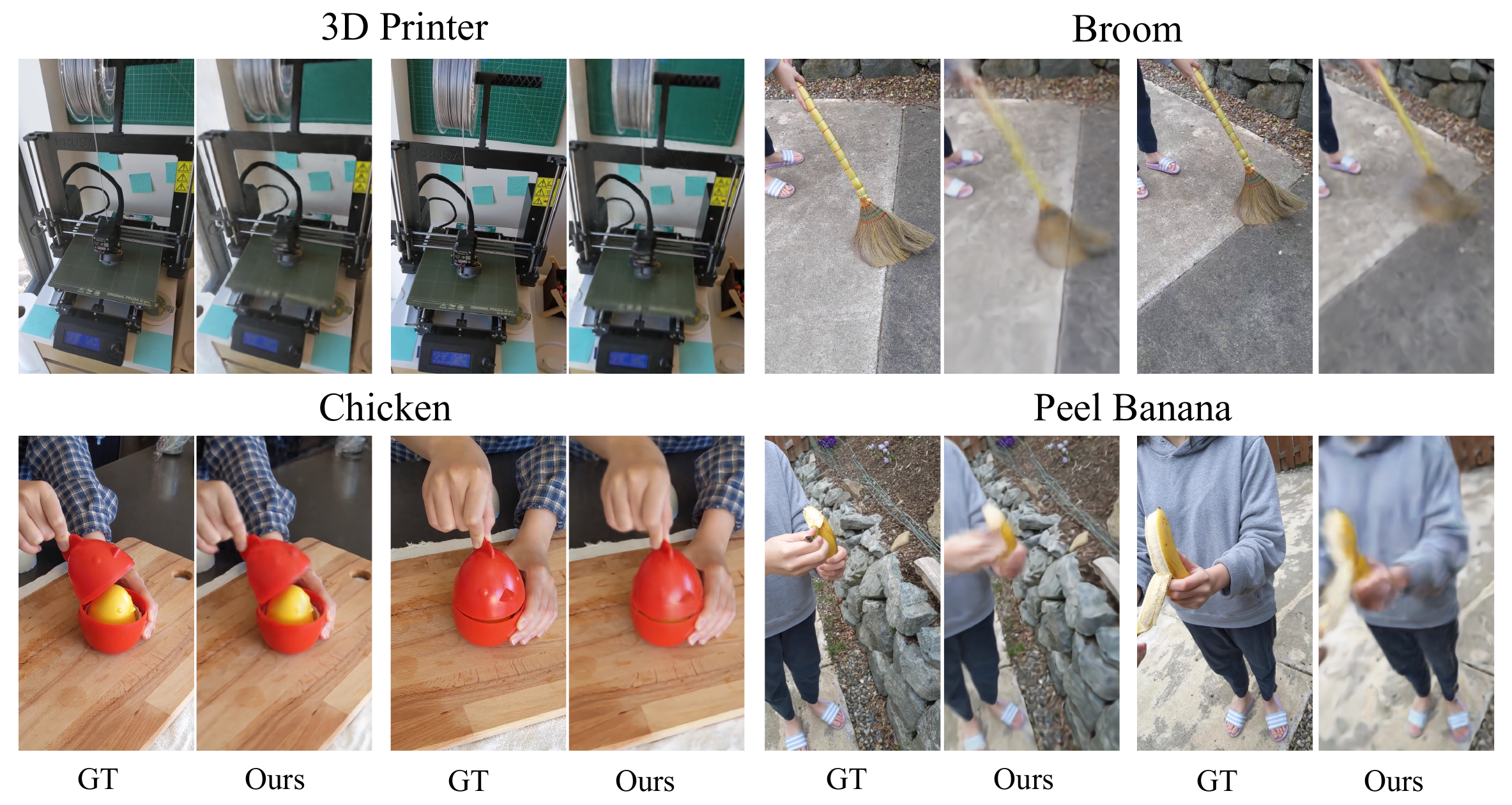}
    \vspace{-0.8em}
    \caption{\footnotesize{Results of real scenes}} \label{fig:hyperimg}
    \vspace{-2.5em}
\end{figure}

First, we show some of the test images of real scenes dataset synthesised by our method in \Fref{fig:hyperimg}. The visual results demonstrate that our model could learn reasonable representation for the real dynamic scenes and enable novel view synthesis with fast optimization.

\begin{figure}[t] \centering
    \vspace{-\baselineskip}
    \includegraphics[width=\textwidth]{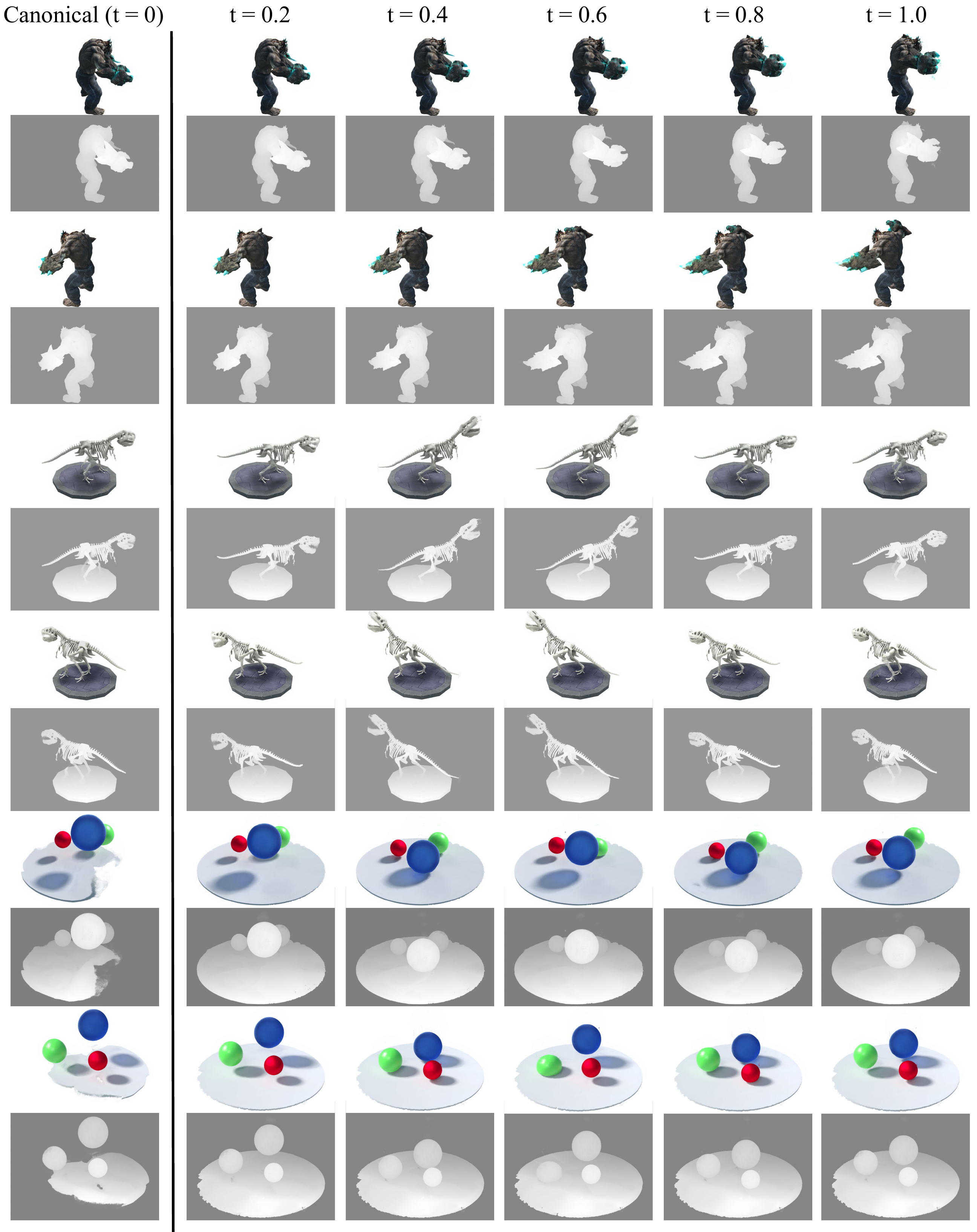}
    \caption{\textbf{More Results for the Learned Geometry.} We show examples of geometries learned by our model. For each, we show rendered images and corresponding disparity under two novel views and six time steps.} \label{fig:suppvisgeo}
\end{figure}

Then, we present more results for the learned geometries in \fref{fig:suppvisgeo}. 
Our method can reconstruct accurate canonical geometry and render high-quality images for different time steps. We notice that the reconstructed canonical geometry of `Bouncing Balls' has some missing parts in the white plate. One of the main reasons is that the background for this scene is white, and the plate has a color similar to the background. The image reconstruction error for this region is low even if the learned density of the plate is close to zero. Then, these low-density regions might be filtered out by our empty space filtering strategy. 
 This problem can be easily alleviated by using a background with different color or disabling the filtering strategy.

\begin{figure}[t] \centering
    \includegraphics[width=\textwidth]{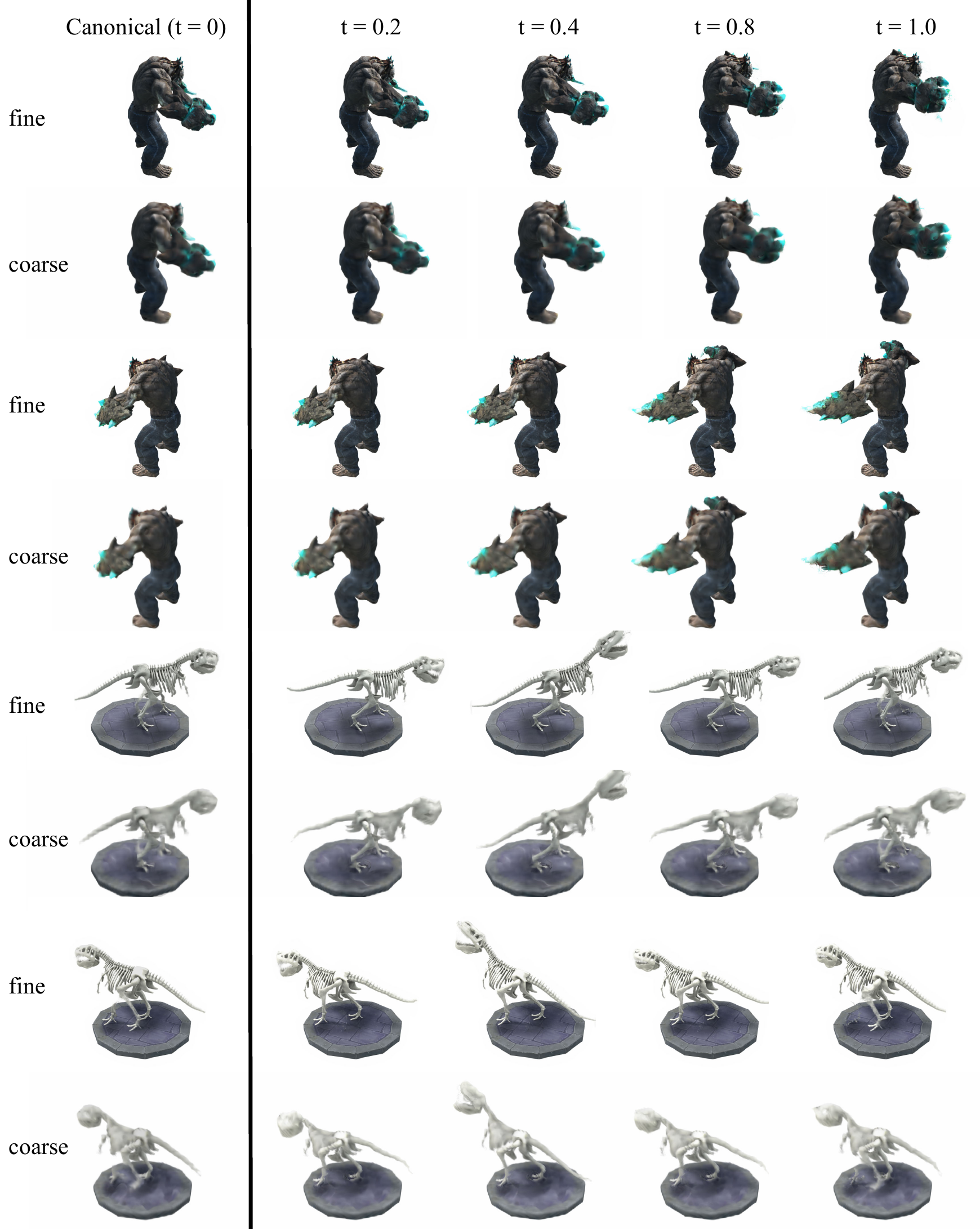}
    \caption{\textbf{Comparison of Results of Coarse and Fine Stage.} Synthesized images rendered by modules trained from coarse stage and fine stage. The images of coarse stage module is relative over smoothed compared with images of fine stage module. Best viewed in color and zoom in for details.} \label{fig:suppcoarsefine}
\end{figure}

\begin{figure}[t] \centering
    \includegraphics[width=\textwidth]{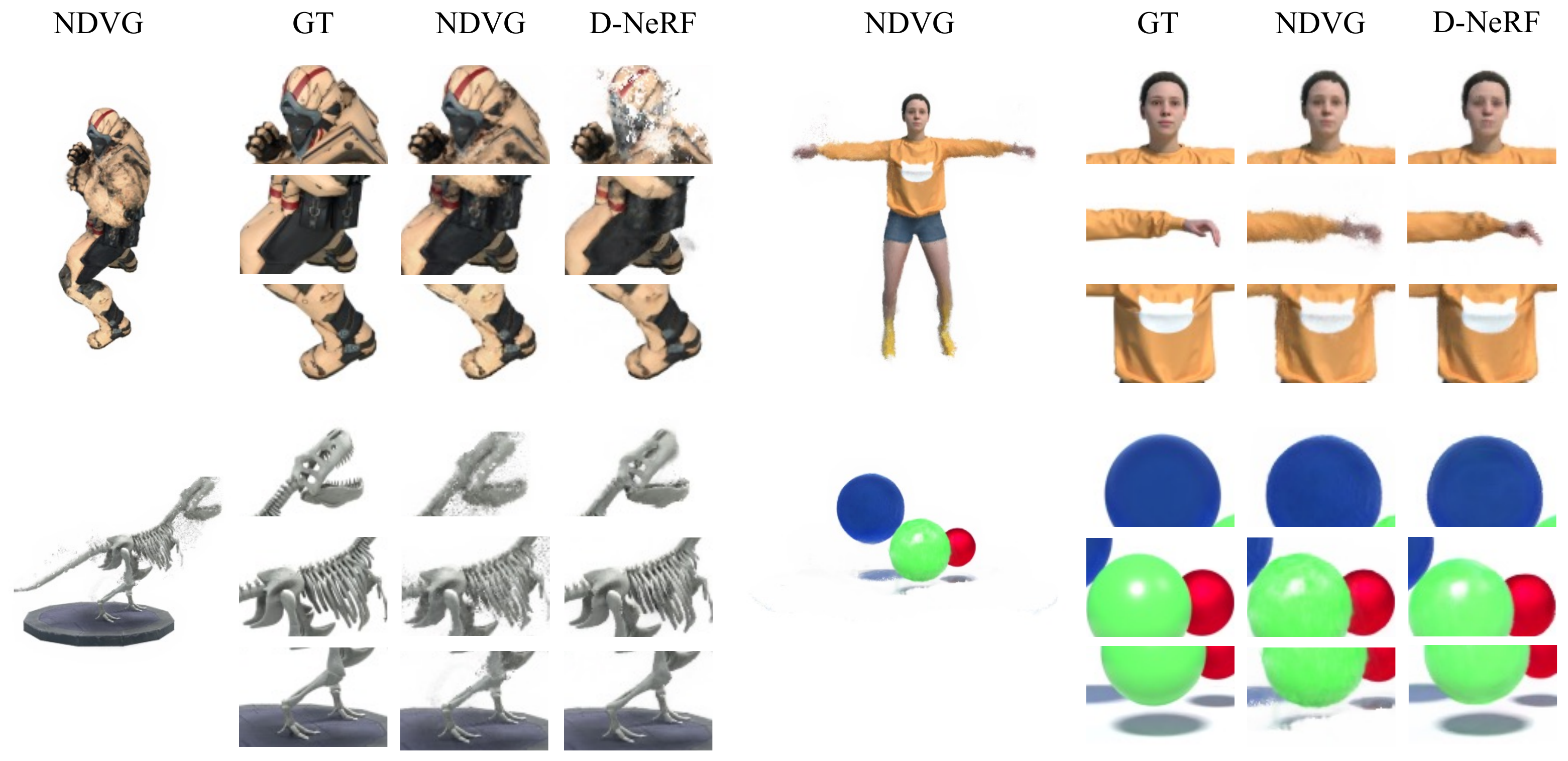}
    \caption{\textbf{More Qualitative Comparison.} Synthesized images on test set of the dataset. For each scene, we show an image rendered at novel view, and followed by zoom in of ground truth, our NDVG, and D-NeRF~\cite{pumarola2021_dnerf_cvpr21}. Best viewed in color and zoom in for details.} \label{fig:suppcompare}
\end{figure}

Finally, we compare the results rendered by the coarse stage and fine stage in \fref{fig:suppcoarsefine}, and we show more results to compare our method with D-NeRF \cite{pumarola2021_dnerf_cvpr21} in \fref{fig:suppcompare}. 

\clearpage

\bibliographystyle{splncs04}
\bibliography{9_References}